\documentclass[10pt,twocolumn,letterpaper]{article}

\usepackage[pagenumbers]{main}

\usepackage{algorithm}
\usepackage{algpseudocode}

\usepackage{colortbl}

\usepackage{fontawesome}

\usepackage{multirow}

\usepackage{titletoc}

\definecolor{ev_1}{RGB}{247,0,9}
\definecolor{ev_2}{RGB}{214,9,42}
\definecolor{ev_3}{RGB}{196,13,61}
\definecolor{ev_4}{RGB}{156,19,102}
\definecolor{ev_5}{RGB}{102,31,157}
\definecolor{ev_6}{RGB}{101,31,158}
\definecolor{ev_7}{RGB}{66,37,194}
\definecolor{ev_8}{RGB}{49,41,211}
\definecolor{ev_9}{RGB}{28,41,206}

\newcommand{\ours}{\textbf{\textcolor{ev_1}{E}\textcolor{ev_2}{-}\textcolor{ev_3}{D}\textcolor{ev_4}{e}\textcolor{ev_5}{f}\textcolor{ev_6}{l}\textcolor{ev_7}{a}\textcolor{ev_8}{r}\textcolor{ev_9}{e}}}

\definecolor{red}{rgb}{0.8,0,0}  
\definecolor{green}{RGB}{0, 133, 21}  
\definecolor{grey}{rgb}{0.5,0.5,0.5}

\definecolor{ev_green}{RGB}{0, 180, 139}
\definecolor{ev_red}{RGB}{239, 99, 75}
\definecolor{ev_blue}{RGB}{99, 113, 250}

\definecolor{cvprblue}{rgb}{0.21,0.49,0.74}
\usepackage[pagebackref=false,breaklinks,colorlinks,allcolors=cvprblue]{hyperref}

\newcommand\blfootnote[1]{%
\begingroup
\renewcommand\thefootnote{}{}\footnote{#1}%
\addtocounter{footnote}{-1}%
\endgroup
}

\makeatletter
\def\blfootnote{\xdef\@thefnmark{}\@footnotetext}
\DeclareRobustCommand\onedot{\futurelet\@let@token\@onedot}
\def\@onedot{\ifx\@let@token.\else.\null\fi\xspace}
\def\eg{\textit{e.g}\onedot}

\def\ie{\textit{i.e}\onedot}

\makeatother

\def\eqref#1{Equation~\ref{#1}}

\usepackage{soul}

\title{Learning to Remove Lens Flare in Event Camera}

\author{
Haiqian Han$^{1,2,3,*}$\quad Lingdong Kong$^{1,2}$\quad Jianing Li$^3$ \quad Ao Liang$^{2}$\quad Chengtao Zhu$^3$\quad Jiacheng Lyu$^3$
\\
Lai Xing Ng$^4$\quad Xiangyang Ji$^3$\quad Wei Tsang Ooi$^2$\quad Benoit R. Cottereau$^{5,6}$
\\[1.2ex]
{\small$^1$CNRS@CREATE}~~
{\small$^2$National University of Singapore}~~
{\small$^3$Tsinghua University}
\\
{\small$^4$Institute for Infocomm Research, A*STAR}~~
{\small$^5$IPAL, CNRS IRL 2955, Singapore}~~
{\small$^6$CerCo, CNRS UMR 5549, Université Toulouse III}
\\[1.2ex]
\faGlobe~\textbf{Project Page:} \href{http://e-flare.github.io/}{\textcolor{ev_blue}{\textbf{\textsl{Link}}}}
~\quad~ 
\faGithubAlt~\textbf{GitHub:} \href{https://github.com/e-flare/toolkit}{\textcolor{ev_red}{\textbf{\textsl{Link}}}}
~\quad~ 
\faDatabase~\textbf{Dataset:} \href{https://huggingface.co/datasets/E-Deflare/data}{\textcolor{ev_green}{\textbf{\textsl{Link}}}}~~
\\[1.8ex]
}

\begin{document}

% Title & Teaser
\twocolumn[{
    \renewcommand\twocolumn[1][]{#1}
    \maketitle
    \begin{center}
    \centering
    \captionsetup{type=figure}
    \vspace{-0.2cm}
    \includegraphics[width=\textwidth]{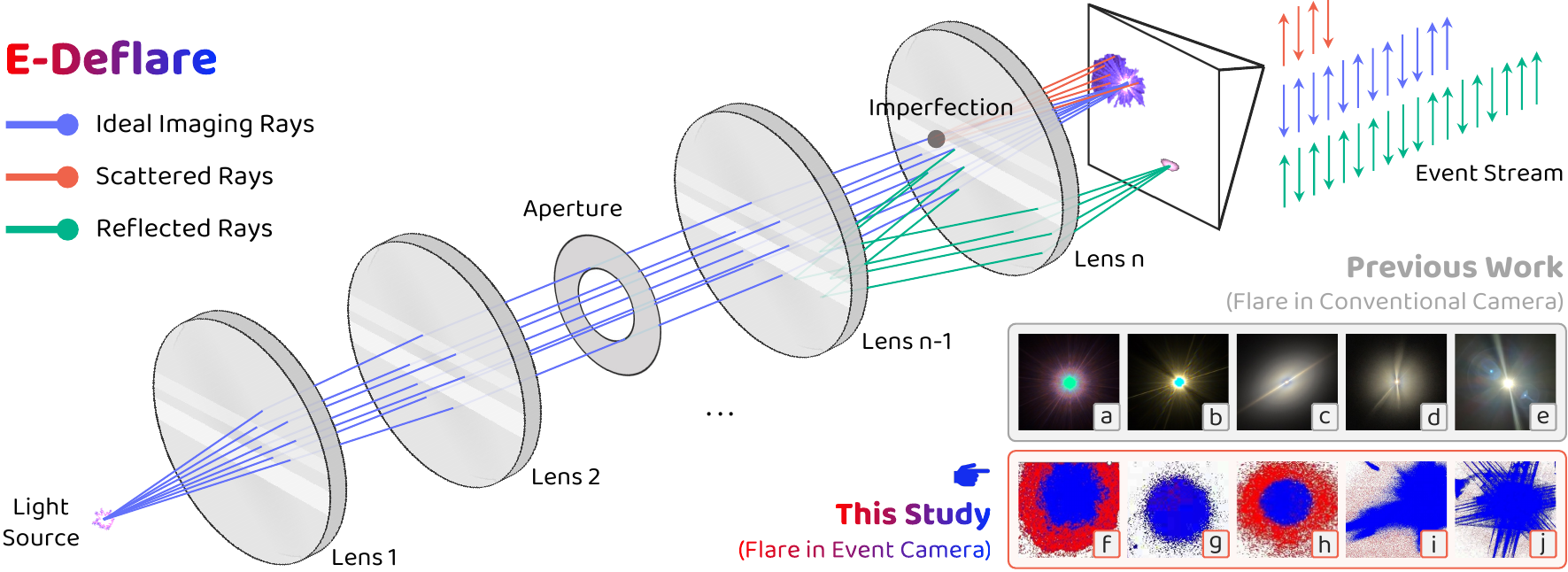}
    \vspace{-0.4cm}
    \caption{\textbf{The physical principle of lens flare and its manifestation in event cameras.} The central diagram deconstructs the universal optical process: ideal \textbf{Imaging Rays} form the scene, while \textbf{Scattered Rays} (from lens imperfections) and \textbf{Internal Reflections} superimpose flare artifacts. While lens flare is a well-studied problem in conventional imaging (\emph{bottom right}, plots $\mathbf{a}$ to $\mathbf{e}$), its complex spatio-temporal manifestation in event data has been largely overlooked (\emph{bottom right}, plots $\mathbf{f}$ to $\mathbf{j}$). Our work is the \textbf{first} to systematically address this challenge. The \textcolor{red}{\textbf{positive}} (brightness increase) and \textcolor{blue}{\textbf{negative}} (brightness decrease) events are colored \textcolor{red}{\textbf{red}} and \textcolor{blue}{\textbf{blue}}, respectively.}
    \label{fig:teaser}
    \vspace{0.6cm}
    \end{center}
}]

% Footnote
\blfootnote{$(*)$ Work done during an internship at CNRS@CREATE.}

% Main Body
\begin{abstract}
Event cameras have the potential to revolutionize vision systems with their high temporal resolution and dynamic range, yet they remain susceptible to lens flare, a fundamental optical artifact that causes severe degradation. In event streams, this optical artifact forms a complex, spatio-temporal distortion that has been largely overlooked. We present \ours, the first systematic framework for removing lens flare from event camera data. We first establish the theoretical foundation by deriving a physics-grounded forward model of the non-linear suppression mechanism. This insight enables the creation of the \textbf{E-Deflare Benchmark}, a comprehensive resource featuring a large-scale simulated training set, \textbf{E-Flare-2.7K}, and the first-ever paired real-world test set, \textbf{E-Flare-R}, captured by our novel optical system. Empowered by this benchmark, we design \textbf{E-DeflareNet}, which achieves state-of-the-art restoration performance. Extensive experiments validate our approach and demonstrate clear benefits for downstream tasks. Code and datasets are publicly available.
\end{abstract}
\vspace{-0.2cm}
\section{Introduction}
\label{sec:intro}

Lens flare \cite{fowles1989introduction,hullin2011physically,wu2021train}, an optical artifact caused by internal reflections and scattering within a camera lens, is a well-known source of image degradation in conventional photography \cite{wu2021train,zhou2023improving,dai2022flare7k,dai2024flare7k++,jiang2024mfdnet}. It severely compromises downstream vision tasks, motivating extensive research on flare removal in the image domain.

Event cameras \cite{gallego2020event,li2021recent,posch2014retinomorphic, chaney2023m3ed,perot2020mpx1,kim2021n-imagenet,muglikar2025event}, which asynchronously capture intensity changes at microsecond resolution, have become increasingly important for high-speed and high-dynamic-range computer vision applications \cite{tulyakov2022time,hagenaars2021self,li2022asynchronous,lin2024e2pnet,vemprala2021representation,wang2023visevent,zheng2023deep,gehrig2024low,han2024event,li2025active,alonso2019ev-segnet,sun2022ess,kong2024openess,kong2025eventfly,kong2025talk2event,kong2025visual,lu2025flexevent,sun2024unified,gehrig2024dense,brander2025reading,messikommer2025data,geckeler2025event}. However, despite their unique sensing paradigm, event cameras are not immune to this fundamental optical limitation \cite{brandli2014davis,posch2010gen1,son2017gen3,finateu2020gen4}. Flare-like artifacts are present in many widely used event datasets~\cite{gehrig2021dsec,gehrig2021raft,hu2020ddd20,binas2017ddd17,zhu2018multivehicle}, where they degrade data quality and hinder downstream performance~\cite{shi2023identifying,shi2024polarity,hidalgo2020learning,zubic2023from,jing2024hpl-ess,hamaguchi2023hmnet}. These artifacts have typically been attributed to generic lighting interference~\cite{shi2023identifying}, without recognizing them as the distinct physical phenomenon of lens flare. We argue that lens flare in event cameras is a critical yet overlooked problem. As shown in Fig.~\ref{fig:teaser}, it manifests as a highly structured, spatio-temporal distortion that can be precisely explained by the underlying physics of optical flare.

This oversight has significant consequences. In event streams, lens flare appears as large-scale, temporally correlated patterns that differ fundamentally from \emph{random noises} or \emph{periodic flickers}. Existing event-based denoising methods \cite{shi2024polarity,shi2023identifying,wang2022linear,guo2022low,duan2021eventzoom}, designed for \emph{uncorrelated} or \emph{uniform noises}, fail to suppress structured artifacts. Moreover, as our theoretical analysis will prove, strong flare signals can suppress or distort valid events from the \emph{background} \cite{valencia2023optically}, making the restoration even more challenging.

A learning-based approach appears most promising, yet it is hindered by a fundamental obstacle: the absence of paired training and testing data~\cite{dai2022flare7k,dai2024flare7k++}. Capturing perfectly aligned event pairs that are both \emph{with} and \emph{without} flare requires re-recording the same dynamic scene under identical motion and illumination, which is infeasible in practice. This challenge is amplified for event cameras, which inherently encode motion over time, making trajectory alignment nearly impossible. Naively applying an event simulator~\cite{hu2021v2e,lin2022dvs,joubert2021event,zhang2023v2ce,han2024physical} to existing image datasets is also inadequate, as it yields a fully synthetic domain for both input and ground truth, resulting in a severe sim-to-real gap.

To overcome the data deadlock, we introduce \emph{\ours}, a novel framework that bridges the gap between optical modeling and data-centric learning. We begin by establishing the theoretical basis with a physical forward model of flare corruption. This model serves as the direct blueprint for our primary contributions: a physics-driven simulator that generates the large-scale \textbf{E-Flare-2.7K} training set, and a novel, physically-principled acquisition system that captures \textbf{E-Flare-R}, the first-ever paired real-world test set. By establishing the first comprehensive benchmark for this task, we can finally train a powerful \textbf{E-DeflareNet} to restore clean event streams with high fidelity.

Our main contributions can be summarized as follows:
\begin{itemize}
    \item We are the first to provide a rigorous physical analysis of lens flare in event data, culminating in a \textbf{forward model} that mathematically formalizes the non-linear suppression mechanism responsible for event stream corruption. This builds a theoretical foundation for de-flaring tasks.

    \item Guided by our physical model, we introduce the first \textbf{simulation pipeline} capable of explicitly modeling the non-linear flare corruption in event streams. This enables, for the first time, the generation of large-scale and physically-grounded paired data for training de-flaring networks.

    \item We design and contribute the \textbf{E-Deflare benchmark}, the first comprehensive resource for this task, comprising a large-scale simulated dataset (\emph{E-Flare-2.7K}), a curated set of real-world examples (\emph{DSEC-Flare}), and critically, the first-ever paired \textbf{real-world test set} (\emph{E-Flare-R}) for comprehensive robustness validations and comparisons.

    \item We design a new restoration model to remove lens flare in event data, achieving state-of-the-art performance compared against strong baselines across all datasets.
\end{itemize}

\section{Related Work}
\label{sec:related_work}

\noindent\textbf{Lens Flare Removal.}
Lens flare is a long-standing problem in image restoration, caused by reflections and scattering within camera lenses \cite{fowles1989introduction,hullin2011physically,wu2021train}. Early heuristic methods \cite{asha2019auto,vitoria2019automatic} have been surpassed by learning-based approaches \cite{dai2023nighttime,dai2022flare7k,wu2021train,dai2024flare7k++,zhou2023improving,jiang2024mfdnet}, which achieve superior performance but face the same challenge: \textit{the lack of paired real-world data} \cite{dai2022flare7k,wu2021train}. This constraint has led to two main directions: unpaired learning with generative models \cite{qiao2021light}, often sacrificing detail, and synthetic training using simulated flare layers \cite{wu2021train,dai2022flare7k,dai2024flare7k++,zhou2023improving}. The success of these simulation-based pipelines highlights their practicality when paired data are unavailable. Yet, all existing methods target static images; they neither address the asynchronous, high-temporal-resolution nature of event data nor the dynamic behavior of flare in this setting.

\noindent\textbf{Event Artifact Removal.}
Event stream restoration has targeted a variety of artifacts. Previous works remove sensor noise such as background activity using hand-crafted filters \cite{shi2024polarity,wu2020probabilistic,guo2022low,feng2020event,wang2019ev,wang2020joint} or learning-based techniques \cite{duan2021eventzoom,rios2023within,baldwin2020event,duan2021guided,zhang2023neuromorphic}. These methods handle random or uncorrelated noise, fundamentally different from the structured patterns caused by lens flare. Others focus on interference from light sources, \eg, flicker removal \cite{wang2022linear,im2023live}, but flare can arise even under stable illumination, making such filters inadequate. A few works mention ``stray light'' artifacts \cite{shi2023identifying,shi2024polarity}, closely related to flare, yet treat them as generic light interference without a physical model. This simplification overlooks the diversity and structure of lens flare, yielding narrow-scope filters that risk suppressing valid events. In contrast, we explicitly model flare as a distinct optical phenomenon and address it with a physics-guided approach.

\noindent\textbf{Event Camera Simulation.}
The scarcity of ground-truth event data has motivated the development of event simulators \cite{bhattacharya2025monocular,messikommer2025data,pellerito2024deep,muglikar2023event}, generally classified into video-driven and learning-based approaches. However, neither can produce paired flare data. The video-driven simulators \cite{rebecq2018esim,hu2021v2e,gehrig2020video,joubert2021event,lin2022dvs,han2024physical} rely on input videos, but paired video flare datasets do not exist, and these methods already suffer from notable sim-to-real gaps \cite{stoffregen2020reducing}, which would worsen when converting synthetic flare videos to synthetic events. Learning-based simulators \cite{zhang2023v2ce,gehrig2020video,pantho2022event,baek2020real,rizzo2022event,zhu2021eventgan,gu2021learn} depend on large and diverse training data that are currently unavailable for flare modeling. Consequently, to the best of our knowledge, \emph{no existing simulator explicitly models the optical process of lens flare in event cameras.} Our work fills this gap by introducing the \textbf{first} physics-based simulator derived from an analytical model of flare formation.

\section{Methodology}
\label{sec:methodology}
This section introduces \emph{\ours}, a physics-guided approach for removing lens flare in event cameras. We begin by deriving a \textbf{forward model} that explains the physical formation of flare and why its removal is an intractable inverse problem (Sec.~\ref{ssec:forward_model}), motivating a learning-based solution. Building on this insight, we develop a two-stage data generation strategy: a \textbf{physics-driven simulator} for large-scale paired training data (Sec.~\ref{ssec:simulator}) and a \textbf{real-world acquisition system} for robust validation (Sec.~\ref{ssec:real_data_acquisition}). With this data infrastructure in place, we define the learning objective and introduce our restoration network.

\begin{figure*}[t]
    \centering
    \includegraphics[width=\textwidth]{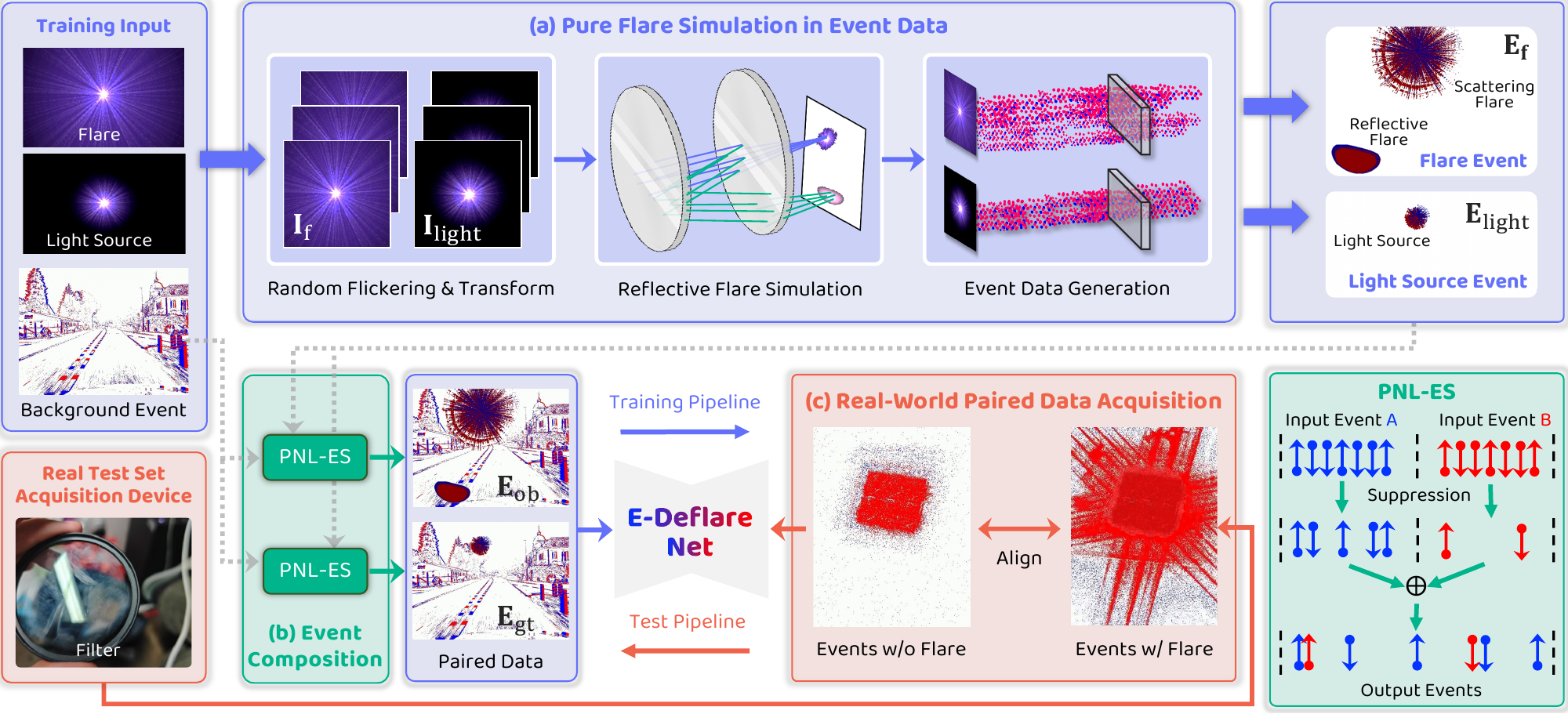}
    \vspace{-0.55cm}
    \caption{\textbf{Data Generation, Training, and Validation.} 
    In the \emph{\ours}~framework, we introduce two parallel pipelines to address data scarcity. 
    The training pipeline synthesizes paired data by first generating dynamic flare events from static images (\eg, Flare7K++) and then fusing them with real background events (\eg, DSEC) via our physics-guided PNL-ES operator. 
    The testing pipeline captures paired real-world data using a controllable optical setup. A removable filter allows us to record the same dynamic scene with flare and without flare, which forms our validation set after post-processing.
    }
    \label{fig:methodology-flowchart}
\end{figure*}

\subsection{Physical Modeling of Flare in Events}
\label{ssec:forward_model}

To understand how lens flare manifests in event streams, we establish a physical model of the \emph{scene} and \emph{imaging process}. The total irradiance captured can be decomposed into distinct components. We define the scene as consisting of a background $\mathbf{I}_{\mathrm{bg}}$ and a primary light source $\mathbf{I}_{\mathrm{light}}$. Under an \textbf{ideal lens} $\mathcal{L}_{\mathrm{ideal}}$, both components are perfectly imaged, producing the clean, flare-free ground-truth irradiance:
\begin{equation}
    \mathbf{I}_{\mathrm{gt}}(x, y, t) = \mathbf{I}_{\mathrm{bg}}(x, y, t) + \mathbf{I}_{\mathrm{light}}(x, y, t)~.
\end{equation}
In contrast, a \textbf{real lens} $\mathcal{L}_{\mathrm{real}}$ introduces internal reflections and scattering that corrupt this formation. While the background remains largely unaffected, the bright light source generates a spatially expansive and structured flare pattern $\mathbf{I}_{\textcolor{ev_red}{\mathrm{\mathbf{f}}}}$, differing substantially from its ideal counterpart $\mathbf{I}_{\mathrm{light}}$. The resulting observed irradiance $\mathbf{I}_{\mathrm{ob}}$ can thus be approximated by a linear superposition in the intensity domain~\cite{dai2022flare7k,wu2021train,dai2024flare7k++}. This process is formulated as follows:
\begin{equation}
    \mathbf{I}_{\mathrm{ob}}(x, y, t) \approx \mathbf{I}_{\mathrm{bg}}(x, y, t) + \mathbf{I}_{\textcolor{ev_red}{\mathrm{\mathbf{f}}}}(x, y, t)~.
\label{eq:image_superposition_final}
\end{equation}
This intensity-domain corruption occurs \emph{before} the sensor and forms the foundation of our event-level analysis. For brevity, we omit coordinates $(x, y)$ where context is clear.

An event camera sensor $\mathcal{S}_{\mathrm{evt}}(\cdot)$ operates on this corrupted irradiance $\mathbf{I}_{\mathrm{ob}}$. Specifically, it asynchronously emits events whenever the change in log-intensity $L(t) = \log(\mathbf{I}_{\mathrm{ob}}(t))$ crosses a contrast threshold $c$. The resulting event stream $\mathbf{E}(t)$ is represented as a sum of Dirac impulses:
\begin{equation}
    \mathbf{E}(t) = \mathcal{S}_{\mathrm{evt}}(\mathbf{I}_{\mathrm{ob}}(t)) = \sum\nolimits_{i} p_i \delta(t - t_i)~,
    \label{eq:event_stream_definition}
\end{equation}
where $p_i \in \{+1, -1\}$ is the event polarity. To characterize how the irradiance superposition in Eq.~\ref{eq:image_superposition_final} propagates into the event domain, we use the integral form of the event camera model, that is:
\begin{equation}
    L(t) = c \int_{0}^{t} \mathbf{E}(\tau) \, d\tau + L(0) + \varepsilon(t)~,
    \label{eq:sensor_model_integral_redef}
\end{equation}
where $L(0)$ is the initial log-intensity and $\varepsilon(t)$ is a bounded quantization error. By differentiating Eq.~\ref{eq:sensor_model_integral_redef} and combining it with Eq.~\ref{eq:image_superposition_final} (see Appendix for derivation), we find that the observed event stream results from a \textbf{non-linear fusion} between background events ($\mathbf{E}_{\mathrm{bg}}$) and flare events ($\mathbf{E}_{\textcolor{ev_red}{\mathrm{\mathbf{f}}}}$). This fusion is dynamically weighted according to the ratio of their linear intensities, that is:
\begin{equation}
    w_{\mathrm{bg}}(t) = \frac{\mathbf{I}_{\mathrm{bg}}(t)}{\mathbf{I}_{\mathrm{bg}}(t) + \mathbf{I}_{\textcolor{ev_red}{\mathrm{\mathbf{f}}}}(t)}~,~~ w_{\textcolor{ev_red}{\mathrm{\mathbf{f}}}}(t) = 1-w_{\mathrm{bg}}(t)~.
    \label{eq:weights_definition}
\end{equation}
In the idealized case, this interaction can be expressed as a weighted sum of event impulses:
\begin{equation}
    \mathbf{E}_{\mathrm{ideal}}(t) = w_{\mathrm{bg}}(t)\mathbf{E}_{\mathrm{bg}}(t) + w_{\textcolor{ev_red}{\mathrm{\mathbf{f}}}}(t)\mathbf{E}_{\textcolor{ev_red}{\mathrm{\mathbf{f}}}}(t)~.
    \label{eq:hypo_stream_final_redef}
\end{equation}
Here, Eq.~\ref{eq:hypo_stream_final_redef} formalizes the \textbf{non-linear suppression effect}: in regions dominated by flare ($\mathbf{I}_{\textcolor{ev_red}{\mathrm{\mathbf{f}}}} \gg \mathbf{I}_{\mathrm{bg}}$), the weight $w_{\mathrm{bg}}$ approaches zero, effectively suppressing genuine background events. However, $\mathbf{E}_{\mathrm{ideal}}(t)$ is a theoretical abstraction; real sensors produce discrete events of fixed amplitude, as defined in Eq.~\ref{eq:event_stream_definition}.

To obtain a physically realizable output, we reconstruct the total linear intensity from $\mathbf{E}_{\mathrm{ideal}}(t)$ and reapply the event generation operator $\mathcal{S}_{\mathrm{evt}}(\cdot)$, that is:
\begin{equation}
    \mathbf{E}_{\mathrm{ob}} = \mathcal{S}_{\mathrm{evt}}\left( \exp\left( c \int_{0}^{t} \mathbf{E}_{\mathrm{ideal}}(\tau) \, d\tau + L(0) \right) \right)~.
    \label{eq:full_forward_model}
\end{equation}
To sum up, Eq.~\ref{eq:full_forward_model} completes our theoretical \textbf{forward model}, revealing that recovering $\mathbf{E}_{\mathrm{gt}}$ from $\mathbf{E}_{\mathrm{ob}}$ is a non-linear and stateful inverse problem that admits no closed-form solution.  This inherent intractability motivates a learning-based restoration approach. While the full physical model is too complex to serve directly as a simulator, the suppression mechanism in Eq.~\ref{eq:hypo_stream_final_redef} captures the essential physics that underpin our data synthesis pipeline. Kindly refer to the Appendix for a more detailed derivation and analysis.

\subsection{Physics-Driven Simulator for Paired Data}
\label{ssec:simulator}

The physical model established in Sec.~\ref{ssec:forward_model} serves as the foundation for our practical simulator. To provide an intuitive understanding of flare artifacts in event data and the quality of our simulation, Fig.~\ref{fig:2x2} presents a qualitative example. It contrasts a real-world flare-corrupted scene with our synthetically generated data pair, demonstrating a high degree of realism. As illustrated in the overview (Fig.~\ref{fig:methodology-flowchart}), the simulator operates in two separate stages to generate the large-scale paired dataset required for training.

\subsubsection{Stage 1: Pure Flare Simulation in Events}
The first stage synthesizes pure event streams for the light source as if captured through both a non-ideal lens ($\mathbf{E}_{\textcolor{ev_red}{\mathrm{\mathbf{f}}}}$) and an ideal lens ($\mathbf{E}_{\mathrm{light}}$). We begin with image assets from Flare7K++ \cite{dai2024flare7k++}, which provide \emph{paired} static flare patterns and clean light sources. However, these assets are inherently static and insufficient for modeling the temporal dynamics of event cameras, as simple image transformations cannot reproduce realistic high-speed variations. 

To address this, we decompose the flare into two components. For the \textbf{scattering flare}, whose pattern varies smoothly with the light source's position, we generate dynamic videos by applying complex temporal effects to the static Flare7K++ assets. These effects include random spatial transformations and intensity flickering to enhance realism. For the  \textbf{reflective flare}, whose intricate patterns change drastically and cannot be realistically interpolated from static images, we design a dedicated \textbf{Reflective Flare Simulation} module to synthesize its dynamic intensity video. The final flare video is the sum of these two components. The resulting synthetic sequences are then processed using a standard event simulator~\cite{lin2022dvs} to produce the pure event streams $\mathbf{E}_{\textcolor{ev_red}{\mathrm{\mathbf{f}}}}$ and $\mathbf{E}_{\mathrm{light}}$.

\subsubsection{Stage 2: Physics-Driven Event Composition}
The second stage composes the final training pairs based on the principles derived in Sec.~\ref{ssec:forward_model}. To model the non-linear interaction between flare and background events, we introduce the \emph{Probabilistic Non-Linear Event Suppression} (\textbf{PNL-ES}) operator, denoted by $\oplus$. This operator merges two event streams according to intensity-dependent probabilistic weights defined in Eq.~\ref{eq:weights_definition}, providing an efficient approximation of the physical suppression mechanism. 

Using this operator, we construct both the flare-corrupted input and its corresponding clean ground truth:
\begin{equation}
    \mathbf{E}_{\mathrm{ob}} = \mathbf{E}_{\mathrm{bg}} \oplus \mathbf{E}_{\textcolor{ev_red}{\mathrm{\mathbf{f}}}} , \quad
\mathbf{E}_{\mathrm{gt}} = \mathbf{E}_{\mathrm{bg}} \oplus \mathbf{E}_{\mathrm{light}}~.
\end{equation}
The required data assets are assembled accordingly. Real-world background events $\mathbf{E}_{\mathrm{bg}}$ are sampled from the DSEC dataset~\cite{gehrig2021dsec}, while the intensity profiles $\mathbf{I}_s(t)$ for the PNL-ES operator are estimated through event accumulation.

This simulation pipeline yields a large-scale, physically grounded dataset for training. By anchoring the synthesis to real background events (which dominate the overall event density), our simulator effectively mitigates the severe sim-to-real gap common to fully synthetic systems. As we will validate in the next sections, this ensures that the resulting network generalizes well to \emph{real-world} data.

\subsection{Real-World Paired Data Acquisition}
\label{ssec:real_data_acquisition}

A key challenge in validation lies in the practical impossibility of capturing \emph{perfectly aligned} dynamic scene pairs with and without flare. We address this by introducing a controllable strategy that records the same underlying scene radiance under two distinct lens configurations.

Our approach is grounded in Fourier optics. The irradiance captured by the sensor in the presence of scattering flare, $\mathbf{I}_{\mathrm{ob}}(\mathbf{t})$, can be expressed as the convolution of the flare-free scene radiance $\mathbf{I}_{\mathrm{gt}}(\mathbf{t})$ (defined in Sec.~\ref{ssec:forward_model}) with the system's Point Spread Function (PSF). The PSF is determined by the lens's pupil function $P(\mathbf{u})$, where $\mathbf{u}$ denotes 2D coordinates in the pupil plane, as $\mathrm{PSF} = |\mathcal{F}\{P(\mathbf{u})\}|^2$~\cite{goodman2005introduction}. To model controllable flare formation, we introduce a multiplicative perturbation $T(\mathbf{u})$ to the clean pupil function $P_{\mathrm{clean}}(\mathbf{u})$, that is:
\begin{equation}
    P_{\mathrm{ob}}(\mathbf{u}) = P_{\mathrm{clean}}(\mathbf{u}) \cdot T(\mathbf{u})~.
\end{equation}
This modification produces two distinct PSFs: a corrupted $\mathrm{PSF}_{\mathrm{ob}}$ (with $T(\mathbf{u}) \neq 1$) and a near-ideal $\mathrm{PSF}_{\mathrm{clean}}$ (with $T(\mathbf{u}) = 1$). Consequently, we can record two event streams of the same dynamic scene, namely the flare-corrupted observation $\mathbf{E}_{\mathrm{ob}}$ and the corresponding near-ideal reference $\tilde{\mathbf{E}}_{\mathrm{gt}}$, which are formulated as follows:
\begin{align}
    \mathbf{E}_{\mathrm{ob}} = \mathcal{S}_{\mathrm{evt}}(\mathbf{I}_{\mathrm{gt}} * \mathrm{PSF}_{\mathrm{ob}})~,~~
    \tilde{\mathbf{E}}_{\mathrm{gt}} = \mathcal{S}_{\mathrm{evt}}(\mathbf{I}_{\mathrm{gt}} * \mathrm{PSF}_{\mathrm{clean}})~.
\end{align}

Physically, we realize $T(\mathbf{u})$ by mounting a removable optical filter in front of the lens (Fig.~\ref{fig:methodology-flowchart}). The capture protocol involves two consecutive recordings of the same dynamic scene: one with the filter in place to obtain $\mathbf{E}_{\mathrm{ob}}$, and another immediately after its removal to acquire $\tilde{\mathbf{E}}_{\mathrm{gt}}$. Multiple filters (\eg, six-line star filters) are used to introduce diversity in flare patterns. The captured stream $\tilde{\mathbf{E}}_{\mathrm{gt}}$ undergoes rigorous post-processing, including residual artifact suppression, facilitated by the known light source geometry in our controlled setup, and metric-guided spatio-temporal alignment to yield the final high-fidelity ground truth $\mathbf{E}_{\mathrm{gt}}$ for validation. This acquisition protocol forms the foundation of our public real-world benchmark.

\begin{figure}[t]
    \centering
    \includegraphics[width=\columnwidth]{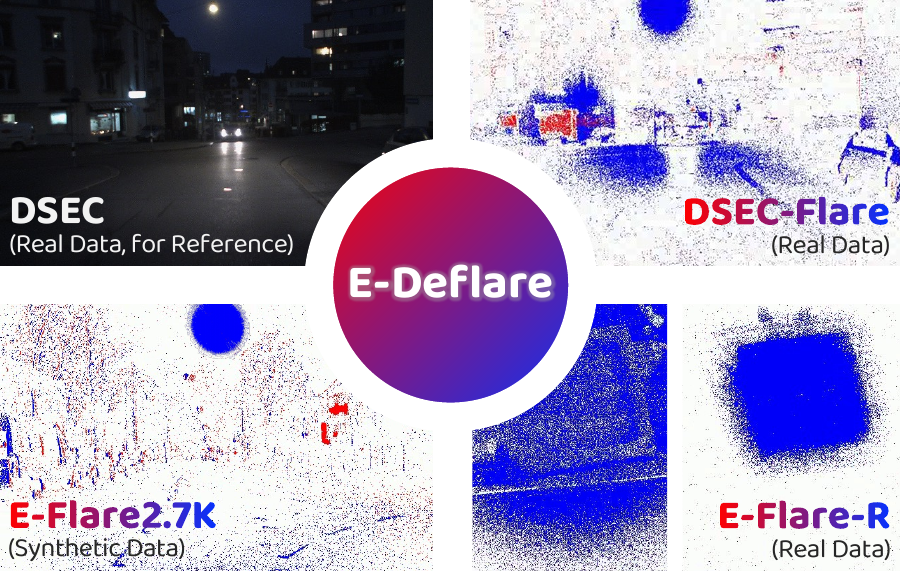}
    \vspace{-0.55cm}
    \caption{Representative data samples from our benchmark, shown alongside a reference RGB frame (top left). The benchmark includes: a real-world flare example from \textbf{DSEC-Flare} (top right); a synthetic sample from \textbf{E-Flare-2.7K} (bottom left); and a paired real-world sample from our test set \textbf{E-Flare-R} (bottom right).}
    \label{fig:2x2}
\end{figure}

With the data generation pipeline complete, we proceed to train our restoration network. The asynchronous event streams are first converted into voxel-grid representations $\mathcal{V}(\cdot)$, a standard structure for event-based processing. We then adopt a 3D U-Net \cite{cciccek20163d}, a widely used architecture for volumetric event representations, and train it in a residual learning manner. The network, termed \textbf{E-DeflareNet}, is trained to predict a residual that approximates the negative flare component. The final restored voxel grid is computed as $\hat{\mathcal{V}}_{\mathrm{gt}} = \mathcal{V}(\mathbf{E}_{\mathrm{ob}}) + f_{\theta}(\mathcal{V}(\mathbf{E}_{\mathrm{ob}}))$. The parameters $\theta$ are optimized via Mean Squared Error (MSE) loss between the restored and ground-truth voxel grids:
\begin{equation}
\theta^* = \underset{\theta}{\arg\min} \left\| \mathcal{V}(\mathbf{E}_{\mathrm{ob}}) + f_{\theta}(\mathcal{V}(\mathbf{E}_{\mathrm{ob}})) - \mathcal{V}(\mathbf{E}_{\mathrm{gt}}) \right\|_2^2.
\end{equation}
Due to space limits, additional architectural and training details of E-DeflareNet are provided in the \textbf{Appendix}.

\subsection{Datasets \& Benchmark}
We establish and contribute the \emph{\ours}~benchmark, comprising three new datasets as follows:
\begin{itemize}
    \item \textbf{\underline{E-Flare2.7K}:} Our large-scale \emph{simulated} event de-flaring dataset (Sec.~\ref{ssec:simulator}). It provides a total of $2.7$K paired samples, where each sample is a $20$ms voxel grid. The dataset is split into $2{,}545$ pairs for training and $175$ for testing.

    \item \textbf{\underline{E-Flare-R}:} Our \emph{real-world} paired test set (Sec.~\ref{ssec:real_data_acquisition}), providing $150$ paired samples for sim-to-real evaluation.

    \item \textbf{\underline{DSEC-Flare}:} A curated set of \emph{real-world} sequences from DSEC~\cite{gehrig2021dsec} to showcase the widespread presence of lens flare in the public and popular event-based dataset.
\end{itemize}
Some data samples are shown in \cref{fig:2x2}. Due to space limits, further details and dataset examples are in the \textbf{Appendix}.
\section{Experiments}
\label{sec:experiments}
This section presents a comprehensive study of event de-flaring through quantitative and qualitative experiments.

\begin{table*}[t]
    \centering
    \caption{\textbf{Quantitative comparisons on the E-Flare2.7K test set}. We evaluate against five event-based de-flaring baselines using event-level and voxel-level metrics, respectively. Important results are color-coded: \textcolor{gray}{Raw Input}, \textcolor{ev_red}{Second-Best}, and \textcolor{ev_blue}{Best} results of under each metric. The final row shows the percentage improvement of our method over the second-best ($\downarrow$: lower is better, $\uparrow$: higher is better).}
    \vspace{-0.3cm}
    \label{tab:main_sim_quant}
    \small
    \begin{tabular*}{\textwidth}{l@{\extracolsep{\fill}}cccccccc}
    \toprule
    \textbf{Method} & Chamfer$\downarrow$ & Gaussian $\downarrow$ & ~MSE $\downarrow$~ & PMSE 2 $\downarrow$ & PMSE 4 $\downarrow$ & ~~R-F1 $\uparrow$~~ & ~~T-F1 $\uparrow$~~ & ~~TP-F1 $\uparrow$~~
    \\
    \midrule
    \textcolor{gray}{$\bullet$}~\textcolor{gray}{Raw Input} & \cellcolor{gray!20}\textcolor{gray}{$1.3555$} & \cellcolor{gray!20}\textcolor{gray}{$0.5222$} & \cellcolor{gray!20}\textcolor{gray}{$0.8315$} & \cellcolor{gray!20}\textcolor{gray}{$0.3666$} & \cellcolor{gray!20}\textcolor{gray}{$0.3430$} & \cellcolor{gray!20}\textcolor{gray}{$0.5138$} & \cellcolor{gray!20}\textcolor{gray}{$0.6303$} & \cellcolor{gray!20}\textcolor{gray}{$0.6939$}
    \\
    \textcolor{ev_red}{$\bullet$}~EFR \cite{wang2022linear} & $1.3237$ & $0.5439$ & $0.5357$ & $0.2019$ & $0.1755$ & $0.1616$ & $0.3572$ & $0.4811$
    \\
    \textcolor{ev_red}{$\bullet$}~PFD-A \cite{shi2024polarity} & $1.3958$ & $0.5397$ & $0.8357$ & $0.3637$ & $0.3392$ & $0.4383$ & \cellcolor{ev_red!20}$0.5496$ & $0.6125$ 
    \\
    \textcolor{ev_red}{$\bullet$}~PFD-B \cite{shi2024polarity} & \cellcolor{ev_red!20}$1.2496$ & \cellcolor{ev_red!20}$0.4613$ & \cellcolor{ev_red!20}$0.2851$ & $0.1159$ & $0.1021$ & \cellcolor{ev_red!20}$0.4460$ & $0.5155$ & $0.5727$ 
    \\
    \textcolor{ev_red}{$\bullet$}~Voxel Transform & $1.2923$ & $0.5576$ & $0.3495$ & \cellcolor{ev_red!20}$0.1037$ & \cellcolor{ev_red!20}$0.0829$ & $0.2177$ & $0.5444$ & \cellcolor{ev_red!20}$0.6318$ 
    \\
    \midrule
    \textcolor{ev_blue}{$\bullet$}~\textbf{E-DeflareNet}~{\textbf{(Ours)}}~~~~~ & \cellcolor{ev_blue!20}$\mathbf{0.4477}$ & \cellcolor{ev_blue!20}$\mathbf{0.2646}$ & \cellcolor{ev_blue!20}$\mathbf{0.1269}$ & \cellcolor{ev_blue!20}$\mathbf{0.0487}$ & \cellcolor{ev_blue!20}$\mathbf{0.0435}$ & \cellcolor{ev_blue!20}$\mathbf{0.7071}$ & \cellcolor{ev_blue!20}$\mathbf{0.7627}$ & \cellcolor{ev_blue!20}$\mathbf{0.7794}$ 
    \\
    ~~~\emph{Improvement (\%)} $\uparrow$ & ~$64.2\%$$\uparrow$ & ~$42.6\%$$\uparrow$ & ~$55.5\%$$\uparrow$ & ~$53.0\%$$\uparrow$ & ~$47.5\%$$\uparrow$ & ~$58.5\%$$\uparrow$ & ~$38.8\%$$\uparrow$ & ~$23.4\%$$\uparrow$
    \\
    \bottomrule
    \end{tabular*}
\end{table*}

\begin{figure*}[t]
    \centering
    \includegraphics[width=\textwidth]{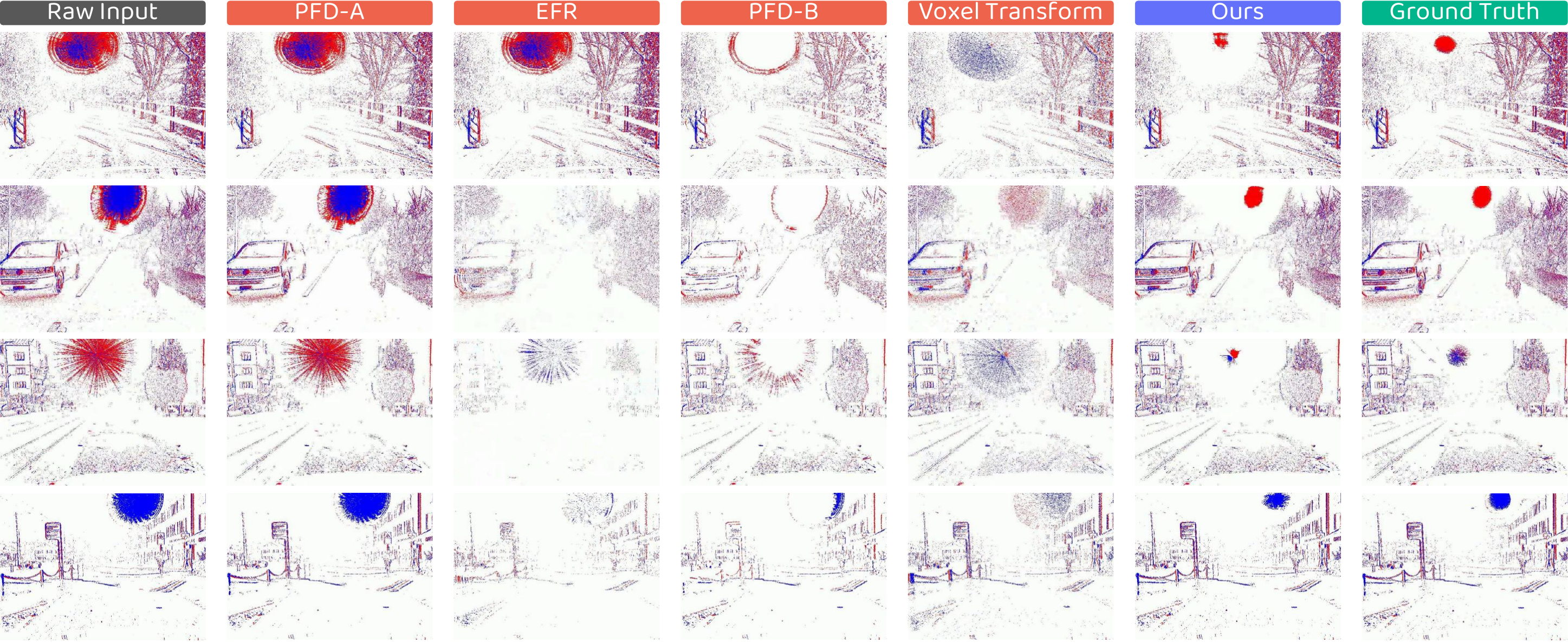}
    \vspace{-0.55cm}
    \caption{\textbf{Qualitative assessments on E-Flare2.7K}. Each column shows the output of a different method applied to the same flare-corrupted input. The rightmost column displays the ground truth for reference. Best viewed in color and zoomed-in for details.}
    \label{fig:qualitative_sim}
\end{figure*}

\subsection{Experimental Settings}

Our model is trained exclusively on \emph{E-Flare2.7K}-train, and tested on \emph{E-Flare2.7K}-test, \emph{E-Flare-R}, and \emph{DSEC-Flare}.

\noindent\textbf{Baselines.} 
We compare against five representative baselines: (1) the unprocessed \emph{RAW Input}; (2) \emph{Voxel Transform}, which evaluates the voxel-grid encoding-decoding effect in isolation; (3) \emph{EFR}~\cite{wang2022linear}, a classical flicker-removal filter; and (4–5) two recent denoising variants from~\cite{shi2024polarity}: \emph{PFD-A} (general noise suppression) and \emph{PFD-B} (light-interference mitigation). More detailed setups are in the \textbf{Appendix}.

\noindent\textbf{Evaluation Protocol.} 
As no established metrics exist for this task, we adapt a set of complementary measures from related event-based studies.  
For \textbf{event-level} fidelity, inspired by PECS~\cite{han2024physical}, we design two metrics: the \emph{Chamfer Distance} ($\downarrow$), computing the average nearest-neighbor distance from the restored events to the ground-truth events; and the \emph{Gaussian Distance} ($\downarrow$), which applies a Gaussian weighting to suppress the influence of outliers. For \textbf{voxel-level} structural quality, we adopt the standard evaluation suite from V2CE~\cite{zhang2023v2ce}, including \emph{Mean Squared Error} (MSE, $\downarrow$), \emph{Pooled MSE} (PMSE, $\downarrow$), and \emph{Binarized-Grid F1} scores (R-F1, T-F1, TP-F1, all $\uparrow$) to jointly capture reconstruction accuracy and precision-recall balance.

\noindent\textbf{Implementation Details.} 
Our model is trained on the \textbf{E-Flare-2.7K} for $40$k iterations, completing within roughly one day on a single NVIDIA RTX $3090$ GPU. The network employs a standard residual 3D U-Net architecture, demonstrating that our physically grounded synthetic data are sufficient for efficient and effective model learning. Kindly refer to the \textbf{Appendix} for more details.

\subsection{Main Results}
\label{sec:main_results}

\noindent\textbf{Evaluation on Simulated Data.}
Table~\ref{tab:main_sim_quant} shows that existing methods are ill-suited for event-based de-flaring. Their rule-based filtering fails to distinguish structured flare patterns from legitimate events, often removing valid signals. In contrast, our learning-based model effectively captures the flare's non-local, spatio-temporal characteristics, enabling it to selectively remove artifacts while preserving scene dynamics. This superiority is highlighted by its commanding performance, achieving, for instance, a $\mathbf{64.2\%}$ improvement in Chamfer distance and a $\mathbf{55.5\%}$ improvement in MSE over the next-best method.

\begin{table*}[t]
    \centering
    \caption{\textbf{Quantitative comparisons on the E-Flare-R dataset}. We evaluate against five event-based de-flaring baselines using event-level and voxel-level metrics, respectively. Important results are color-coded: \textcolor{gray}{Raw Input}, \textcolor{ev_red}{Second-Best}, and \textcolor{ev_blue}{Best} results of under each metric. The final row shows the percentage improvement of our method over the second-best ($\downarrow$: lower is better, $\uparrow$: higher is better).}
    \vspace{-0.3cm}
    \label{tab:main_real_quant}
    \small
    \begin{tabular*}{\textwidth}{l@{\extracolsep{\fill}}cccccccc}
    \toprule
    \textbf{Method} & Chamfer$\downarrow$ & Gaussian $\downarrow$ & ~MSE $\downarrow$~ & PMSE 2 $\downarrow$ & PMSE 4 $\downarrow$ & ~~R-F1 $\uparrow$~~ & ~~T-F1 $\uparrow$~~ & ~~TP-F1 $\uparrow$~~
    \\
    \midrule
    \textcolor{gray}{$\bullet$}~\textcolor{gray}{Raw Input} & \cellcolor{gray!20}\textcolor{gray}{$1.7855$} & \cellcolor{gray!20}\textcolor{gray}{$0.7170$} & \cellcolor{gray!20}\textcolor{gray}{$0.8158$} & \cellcolor{gray!20}\textcolor{gray}{$0.3431$} & \cellcolor{gray!20}\textcolor{gray}{$0.3266$} & \cellcolor{gray!20}\textcolor{gray}{$0.2299$} & \cellcolor{gray!20}\textcolor{gray}{$0.2915$} & \cellcolor{gray!20}\textcolor{gray}{$0.3093$}
    \\
    \textcolor{ev_red}{$\bullet$}~EFR \cite{wang2022linear} & $1.8191$ & $0.7266$ & $0.3388$ & $0.1403$ & $0.1330$ & $0.0839$ & $0.1807$ & $0.2372$ 
    \\
    \textcolor{ev_red}{$\bullet$}~PFD-A \cite{shi2024polarity} & \cellcolor{ev_red!20}$1.7642$ & \cellcolor{ev_red!20}$0.7067$ & $0.7924$ & $0.3341$ & $0.3182$ & \cellcolor{ev_red!20}$0.2386$ & \cellcolor{ev_red!20}$0.3081$ & \cellcolor{ev_red!20}$0.3283$
    \\
    \textcolor{ev_red}{$\bullet$}~PFD-B \cite{shi2024polarity} & $2.0838$ & $0.8504$ & \cellcolor{ev_red!20}$0.2759$ & $0.1121$ & $0.1049 $& $0.0300$ & $0.0717$ & $0.0880$ 
    \\
    \textcolor{ev_red}{$\bullet$}~Voxel Transform & $1.9535$ & $0.8167$ & $0.3140$ & \cellcolor{ev_red!20}$0.0964$ & \cellcolor{ev_red!20}$0.0836$ & $0.1036$ & $0.2729$ & $0.3255$ 
    \\
    \midrule
    \textcolor{ev_blue}{$\bullet$}~\textbf{E-DeflareNet}~{\textbf{(Ours)}}~~~~~ & \cellcolor{ev_blue!20}$\mathbf{1.1368}$ & \cellcolor{ev_blue!20}$\mathbf{0.4651}$ & \cellcolor{ev_blue!20}$\mathbf{0.1741}$ & \cellcolor{ev_blue!20}$\mathbf{0.0690}$ & \cellcolor{ev_blue!20}$\mathbf{0.0644}$ & \cellcolor{ev_blue!20}$\mathbf{0.3498}$ & \cellcolor{ev_blue!20}$\mathbf{0.3386}$ & \cellcolor{ev_blue!20}$\mathbf{0.3011}$
    \\
    ~~~\emph{Improvement (\%)} $\uparrow$ & ~$35.6\%$$\uparrow$ & ~$34.2\%$$\uparrow$ & ~$36.9\%$$\uparrow$ & ~$28.4\%$$\uparrow$ & ~$23.0\%$$\uparrow$ & ~$46.6\%$$\uparrow$ & ~$9.9\%$$\uparrow$ & ~-$8.3\%$$\downarrow$ 
    \\
    \bottomrule
    \end{tabular*}
\end{table*}

\begin{figure*}[t]
    \centering
    \includegraphics[width=\textwidth]{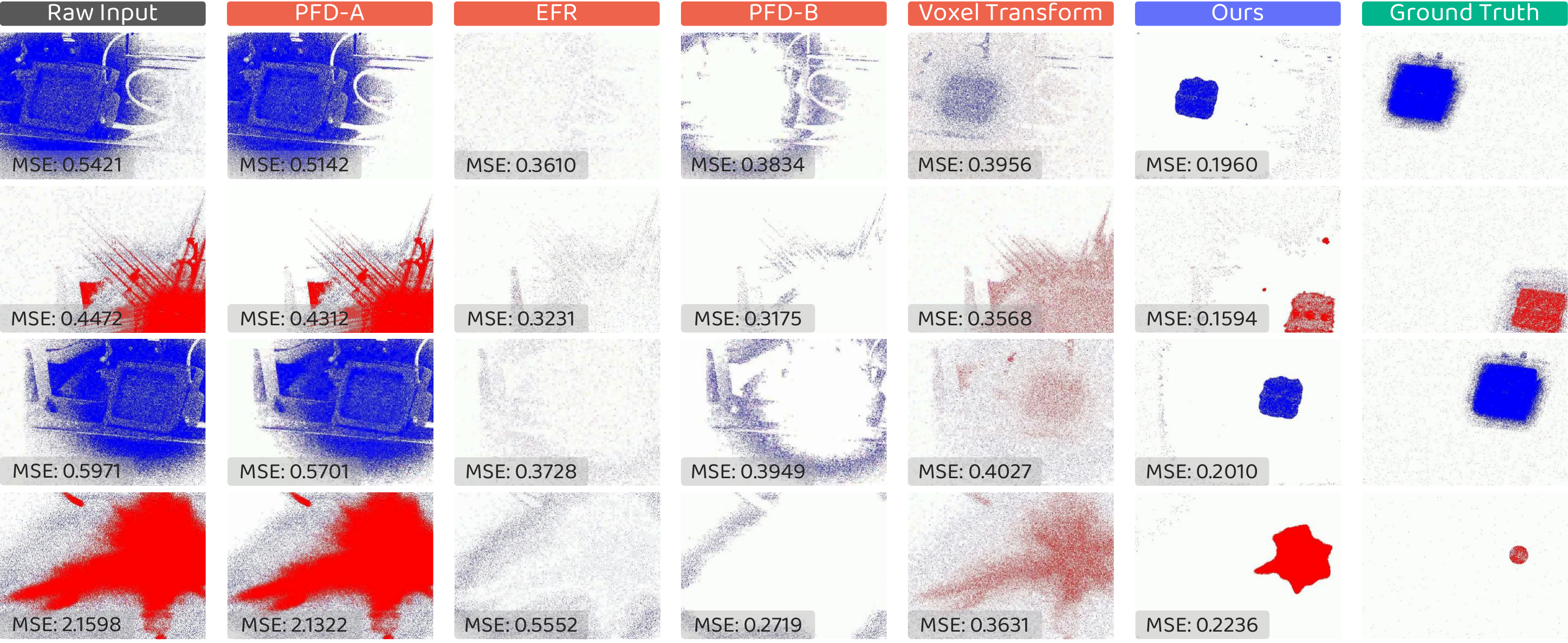}
    \vspace{-0.55cm}
    \caption{\textbf{Qualitative assessments on E-Flare-R.}  Each column shows the output of a different method applied to the same flare-corrupted input. The rightmost column displays the ground truth for reference. Best viewed in color and zoomed-in for details.}
    \label{fig:qualitative_real}
\end{figure*}

\begin{table*}[t]
    \centering
    \caption{\textbf{Ablation study on the E-Flare-R dataset}. We evaluate five variants to analyze our design choices. The final row shows the improvement of our full model over the best ablated variant ($\downarrow$: lower is better, $\uparrow$: higher is better).}
    \vspace{-0.3cm}
    \label{tab:ablation_main}
    \small
    \begin{tabular*}{\textwidth}{l@{\extracolsep{\fill}}cccccccc}
    \toprule
    \textbf{Configuration} & Chamfer$\downarrow$ & Gaussian $\downarrow$ & ~MSE $\downarrow$~ & PMSE 2 $\downarrow$ & PMSE 4 $\downarrow$ & ~~R-F1 $\uparrow$~~ & ~~T-F1 $\uparrow$~~ & ~~TP-F1 $\uparrow$~~
    \\
    \midrule
    \textcolor{gray}{$\bullet$}~\textcolor{gray}{Raw Input} & \cellcolor{gray!20}\textcolor{gray}{$1.7855$} & \cellcolor{gray!20}\textcolor{gray}{$0.7170$} & \cellcolor{gray!20}\textcolor{gray}{$0.8158$} & \cellcolor{gray!20}\textcolor{gray}{$0.3431$} & \cellcolor{gray!20}\textcolor{gray}{$0.3266$} & \cellcolor{gray!20}\textcolor{gray}{$0.2299$} & \cellcolor{gray!20}\textcolor{gray}{$0.2915$} & \cellcolor{gray!20}\textcolor{gray}{$0.3093$}
    \\\midrule
    \textcolor{ev_red}{$\bullet$}~Simple & \cellcolor{ev_red!20}$1.3434$ & \cellcolor{ev_red!20}$0.5564$ & \cellcolor{ev_red!20}$0.1816$ & \cellcolor{ev_red!20}$0.0734$ & \cellcolor{ev_red!20}$0.0688$ & \cellcolor{ev_red!20}$0.2703$ & $0.2768$ & $0.2644$ 
    \\
    \textcolor{ev_red}{$\bullet$}~Simple $+$ Random & $1.5257$ & $0.6449$ & $0.1957$ & $0.0769$ & $0.0715$ & $0.2301$ & \cellcolor{ev_red!20}$0.2769$ & \cellcolor{ev_red!20}$0.2811$ 
    \\
    \textcolor{ev_red}{$\bullet$}~Physical $+$ Random & $1.6811$ & $0.7203$ & $0.2020$ & $0.0834$ & $0.0788$ & $0.1567$ & $0.2001$ & $0.2103$ 
    \\
    \textcolor{ev_green}{$\bullet$}~No Light & $1.9745$ & $0.8487$ & $0.1951$ & $0.0843$ & $0.0806$ & $0.0189$ & $0.0459$ & $0.0709$ 
    \\
    \midrule
    \textcolor{ev_blue}{$\bullet$}~\textbf{Full Configuration}~~~~~ & \cellcolor{ev_blue!20}$\mathbf{1.1368}$ & \cellcolor{ev_blue!20}$\mathbf{0.4651}$ & \cellcolor{ev_blue!20}$\mathbf{0.1741}$ & \cellcolor{ev_blue!20}$\mathbf{0.0690}$ & \cellcolor{ev_blue!20}$\mathbf{0.0644}$ & \cellcolor{ev_blue!20}$\mathbf{0.3498}$ & \cellcolor{ev_blue!20}$\mathbf{0.3386}$ & \cellcolor{ev_blue!20}$\mathbf{0.3011}$
    \\
    ~~~\emph{Improvement (\%)} $\uparrow$ & ~$15.4\%$$\uparrow$ & ~$16.4\%$$\uparrow$ & ~$4.1\%$$\uparrow$ & ~$6.0\%$$\uparrow$ & ~$6.4\%$$\uparrow$ & ~$29.4\%$$\uparrow$ & ~$22.3\%$$\uparrow$ & ~$7.1\%$$\uparrow$
    \\
    \bottomrule
    \end{tabular*}
\end{table*}

\begin{table}[t]
    \centering
    \caption{\textbf{Downstream task comparisons} for event-based 3D reconstruction. We use Event3DGS~\cite{han2024event} to reconstruct scenes and evaluate the quality of novel view synthesis. $\uparrow$: higher is better.}
    \vspace{-0.3cm}
    \label{tab:3dgs_quant}
    \small
    \begin{tabular*}{\columnwidth}{r|c|ccc|c}
    \toprule
    \textbf{Metric} & \textcolor{gray}{Original} & EFR & PFD-A & PFD-B & \textbf{Ours} 
    \\
    \midrule
    PSNR $\uparrow$ & \cellcolor{gray!20}\textcolor{gray}{$13.72$} & \cellcolor{ev_red!20}$13.74$ & $13.11$ & $13.32$ & \cellcolor{ev_blue!20}$\mathbf{13.78}$
    \\
    SSIM $\uparrow$ & \cellcolor{gray!20}\textcolor{gray}{$0.765$} & $0.741$ & \cellcolor{ev_red!20}$0.743$ & $0.734$ & \cellcolor{ev_blue!20}$\mathbf{0.792}$
    \\
    \bottomrule
    \end{tabular*}
    \vspace{0.1cm}
\end{table}

\noindent\textbf{Evaluation on Paired Real-World Data.}
To assess real-world generalization, we perform a zero-shot sim-to-real evaluation, as our model is trained \emph{solely} on simulated data. As shown in Table~\ref{tab:main_real_quant}, the performance trends remain largely consistent. \textbf{E-DeflareNet} significantly outperforms all baselines on most key metrics, such as achieving a $\mathbf{35.6\%}$ improvement in Chamfer distance over the second-best method. While TP-F1 shows a slight decrease (-$8.3\%$), this reflects a conservative strategy that prioritizes high fidelity by avoiding false positives over aggressively recovering every noisy event. This trade-off leads to superior results on stricter metrics like Chamfer and R-F1, which penalize structural distortions. The strong overall performance validates both the robustness of our model and the effectiveness of our simulation pipeline.

\noindent\textbf{Qualitative Analysis.}
The visual comparisons in Fig.~\ref{fig:qualitative_sim} and Fig.~\ref{fig:qualitative_real} further illustrate these findings. Baselines tend to leave substantial flare artifacts or excessively suppress light sources, compromising scene integrity. In contrast, \textbf{E-DeflareNet} exhibits precise spatial discrimination, cleanly removing the structured flare while preserving the underlying illumination consistent with the ground truth, thereby recovering physically plausible, flare-free event streams.

\subsection{Ablation Study}
\label{sec:ablation}

We conduct ablation studies on the E-Flare-R to examine the impact of our core design choices (Table~\ref{tab:ablation_main}). Five variants are evaluated:  
\textbf{(1)} \emph{`Model No Light'} (trained with flawed ground truth that omits the valid light source);  
\textbf{(2)} \emph{`Simple Model'} (direct additive fusion of event streams);  
\textbf{(3)} \emph{`Simple Model + Random'} (the same model with random temporal jitter);  
\textbf{(4)} \emph{`Physical Model'} (our proposed probabilistic gating mechanism); and  
\textbf{(5)} \emph{`Physical Model + Random'} (the physical model with temporal jitter applied).

The results reveal a clear hierarchy of performance and validate our design rationale.  
First, the drastic degradation of \emph{`Model No Light'} confirms that preserving the true light source is essential, as treating de-flaring as pure removal leads to severe loss of semantic content. Second, the large gap between \emph{`Simple Model'} and \emph{`Physical Model'} underscores the necessity of our physics-guided formulation. To ensure this improvement is not merely due to stochastic augmentation, we compare with \emph{`Simple Model + Random'}, which remains far below the physical variant, verifying that our gains stem from the probabilistic gating mechanism rather than randomness. Finally, the drop in \emph{`Physical Model + Random'} highlights the importance of precise, physically grounded suppression: disrupting its temporal consistency directly harms performance.  

Collectively, we confirm that the success of \textbf{E-Deflare} arises from the principled non-linear suppression model that faithfully captures the physics of flare corruption.

\begin{figure}[t]
    \centering
    \includegraphics[width=\columnwidth]{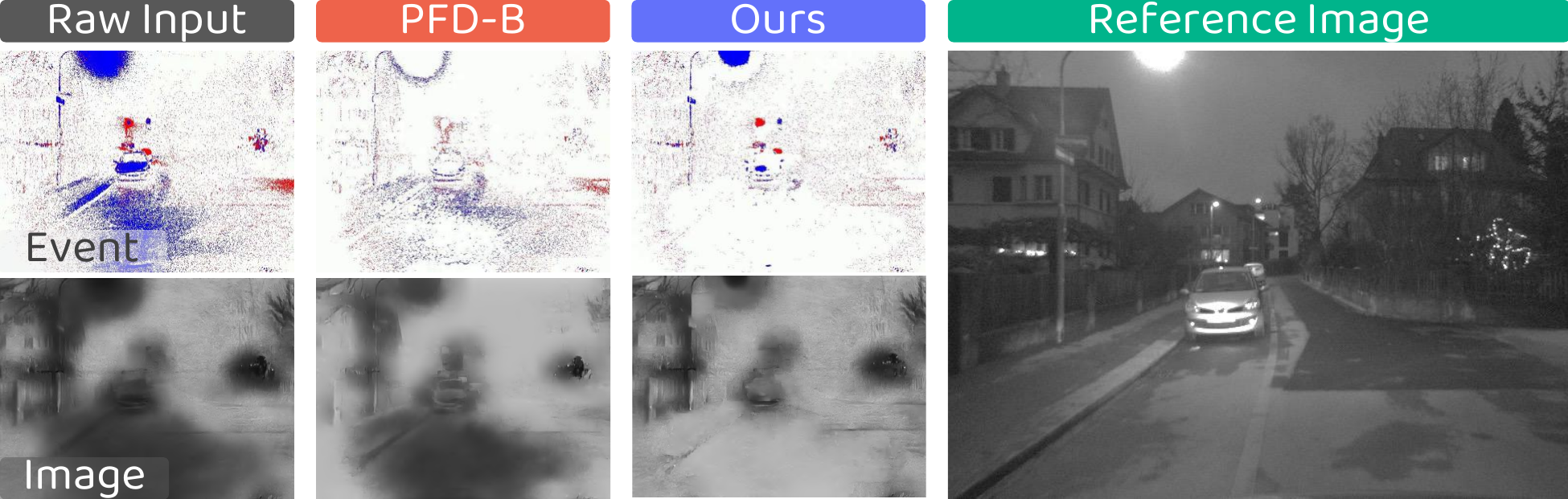}
    \vspace{-0.65cm}
    \caption{\textbf{Downstream task comparisons} for event-based imaging. We use SPADE-E2VID~\cite{cadena2021spade} to reconstruct images from both raw and our de-flared event streams on a challenging DSEC scene.}
    \label{fig:image_reconstruction}
\end{figure}

\begin{figure}[t]
    \centering
    \includegraphics[width=\columnwidth]{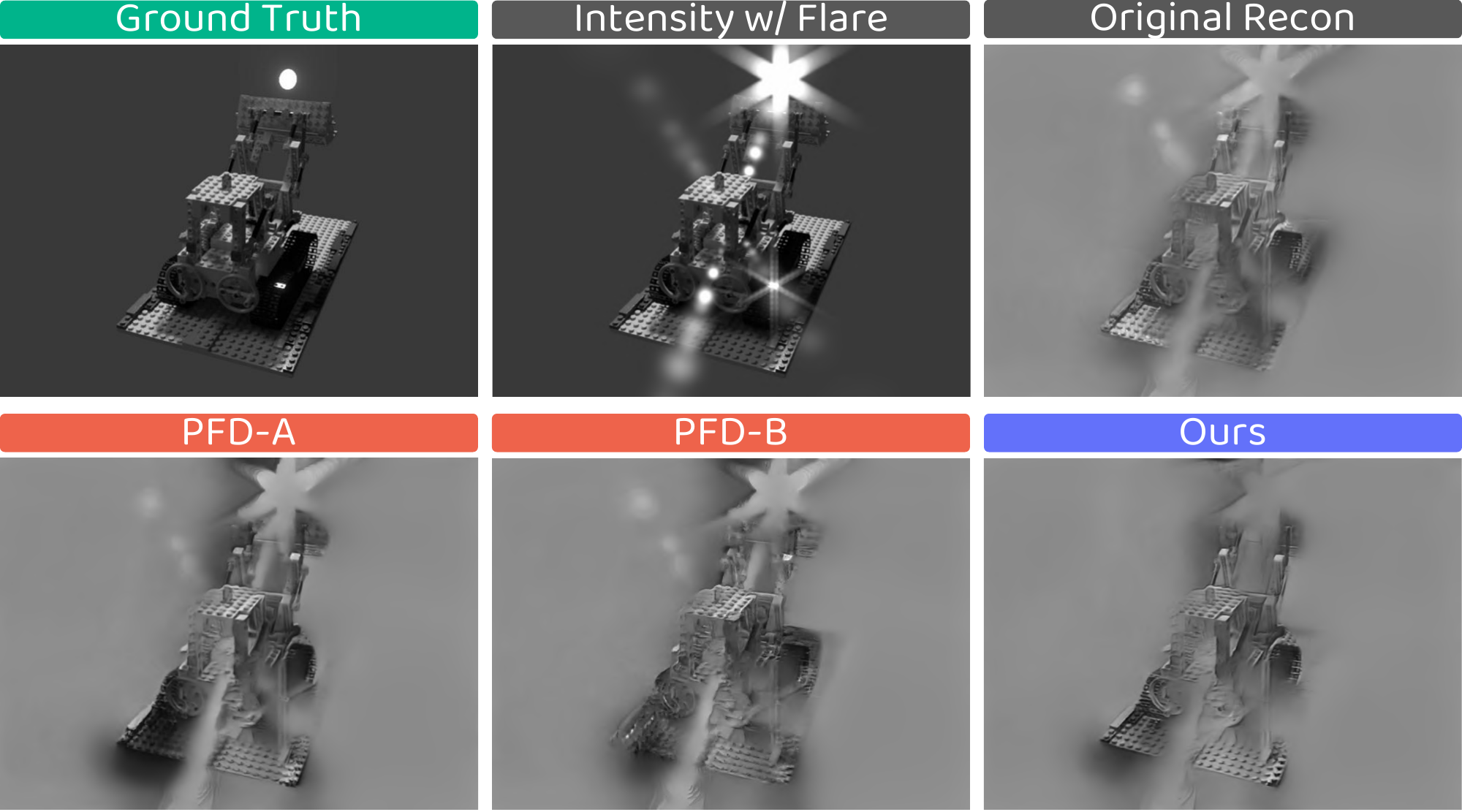}
    \vspace{-0.65cm}
    \caption{\textbf{Downstream task comparisons} for event-based 3D reconstruction. We visualize the rendered views from 3D scenes reconstructed using the outputs of different de-flaring methods.}
    \label{fig:3dgs_qual}
\end{figure}

\subsection{Downstream Task Evaluations}
\label{sec:downstream}

To demonstrate the practical benefits of \textbf{E-Deflare}, we evaluate its effect on two high-level event-based vision tasks: image reconstruction and 3D scene reconstruction.

\noindent\textbf{Event-based Imaging.}
We first assess the impact of de-flaring on event-to-image reconstruction using the state-of-the-art SPADE-E2VID framework~\cite{cadena2021spade}. Fig.~\ref{fig:image_reconstruction} shows results on challenging sequences in \textbf{DSEC-Flare}. Images reconstructed from raw, flare-corrupted events exhibit strong flare patterns, obscuring fine scene details. In contrast, images reconstructed from our de-flared events recover cleaner and accurate illumination, revealing significant visual improvement. These results demonstrate that \textbf{E-Deflare} serves as an effective pre-processing step, substantially enhancing the clarity and usability of event data for imaging applications.

\noindent\textbf{Event-based 3D Reconstruction.}
To demonstrate the practical impact of our method, we conduct a rigorous quantitative evaluation on event-based 3D reconstruction. As no standard benchmark with flare artifacts exists for this task, we synthesized a paired event dataset from the popular LEGO scene of NeRF-synthetic~\cite{mildenhall2021nerf}, following the simulation methodology of PECS~\cite{han2024physical}.
The experimental process involves reconstructing a 3D scene from various event streams (the original flare-corrupted input, our de-flared output, and outputs from baseline methods) using the method from Event3DGS~\cite{han2024event}. We then evaluate the reconstruction by rendering novel views and comparing them against the pristine ground truth, with background color normalization for a fair comparison.

The results in Table~\ref{tab:3dgs_quant} reveals that naive de-flaring can be detrimental. Most baselines fail to improve reconstruction, with some even degrading performance below the original input. In contrast, our method is the only one providing a clear and consistent improvement, boosting the SSIM from $0.765$ to a state-of-the-art $\mathbf{0.792}$. This powerfully demonstrates that our approach provides a cleaner, more reliable input that directly benefits downstream 3D perception.
\section{Conclusion}
\label{sec:conclusion}

We introduced \textbf{E-Deflare}, the first systematic framework to address the critical and overlooked problem of lens flare in event camera data. We established the theoretical foundation with a physics-based forward model of the non-linear flare corruption. This model guided the development of two core contributions: a physics-driven simulator that produced our large-scale \textbf{E-Flare-2.7K} training set, and a novel, Fourier-optics-based acquisition system that captured \textbf{E-Flare-R}, the first paired real-world test set. The combination of these innovations forms the \textbf{E-Deflare Benchmark}, a principled resource that enables our \textbf{E-DeflareNet} to achieve state-of-the-art performance and markedly improve downstream tasks. Beyond providing a robust de-flaring solution, our work offers a valuable blueprint for tackling other data-scarce challenges in event vision.
Future efforts will explore more powerful architectures and multi-modal fusion with frame-based sensors.

\section*{Acknowledgments}
This work is under the programme DesCartes and is supported by the National Research Foundation, Prime Minister’s Office, Singapore, under its Campus for Research Excellence and Technological Enterprise (CREATE) programme. This work is also supported by the Apple Scholars in AI/ML Ph.D. Fellowship program.

{
    \small
    \bibliographystyle{ieeenat_fullname}
    \bibliography{main}

\begin{thebibliography}{89}
\providecommand{\natexlab}[1]{#1}
\providecommand{\url}[1]{\texttt{#1}}
\expandafter\ifx\csname urlstyle\endcsname\relax
  \providecommand{\doi}[1]{doi: #1}\else
  \providecommand{\doi}{doi: \begingroup \urlstyle{rm}\Url}\fi

\bibitem[Alonso and Murillo(2019)]{alonso2019ev-segnet}
Inigo Alonso and Ana~C. Murillo.
\newblock {EV-SegNet}: Semantic segmentation for event-based cameras.
\newblock In \emph{IEEE/CVF Conference on Computer Vision and Pattern Recognition Workshops}, pages 1--10, 2019.

\bibitem[Asha et~al.(2019)Asha, Bhat, Nayak, and Bhat]{asha2019auto}
CS Asha, Sooraj~Kumar Bhat, Deepa Nayak, and Chaithra Bhat.
\newblock Auto removal of bright spot from images captured against flashing light source.
\newblock In \emph{IEEE International Conference on Distributed Computing, VLSI, Electrical Circuits and Robotics}, pages 1--6, 2019.

\bibitem[Baek et~al.(2020)Baek, Eshraghian, Thio, Sandamirskaya, Iu, and Lu]{baek2020real}
Seungbum Baek, Jason~K Eshraghian, Wesley Thio, Yulia Sandamirskaya, Herbert~HC Iu, and Wei~D Lu.
\newblock A real-time retinomorphic simulator using a conductance-based discrete neuronal network.
\newblock In \emph{IEEE International Conference on Artificial Intelligence Circuits and Systems}, pages 79--83, 2020.

\bibitem[Baldwin et~al.(2020)Baldwin, Almatrafi, Asari, and Hirakawa]{baldwin2020event}
R Baldwin, Mohammed Almatrafi, Vijayan Asari, and Keigo Hirakawa.
\newblock Event probability mask ({EPM}) and event denoising convolutional neural network ({EDNCNN}) for neuromorphic cameras.
\newblock In \emph{IEEE/CVF Conference on Computer Vision and Pattern Recognition}, pages 1701--1710, 2020.

\bibitem[Bhattacharya et~al.(2025)Bhattacharya, Cannici, Rao, Tao, Kumar, Matni, and Scaramuzza]{bhattacharya2025monocular}
Anish Bhattacharya, Marco Cannici, Nishanth Rao, Yuezhan Tao, Vijay Kumar, Nikolai Matni, and Davide Scaramuzza.
\newblock Monocular event-based vision for obstacle avoidance with a quadrotor.
\newblock In \emph{Conference on Robot Learning}, pages 4826--4843. PMLR, 2025.

\bibitem[Binas et~al.(2017)Binas, Neil, Liu, and Delbruck]{binas2017ddd17}
Jonathan Binas, Daniel Neil, Shih-Chii Liu, and Tobi Delbruck.
\newblock {DDD17}: End-to-end {DAVIS} driving dataset.
\newblock In \emph{International Conference on Machine Learning Workshops}, pages 1--9, 2017.

\bibitem[Brander et~al.(2025)Brander, Cioffi, Messikommer, and Scaramuzza]{brander2025reading}
Carl Brander, Giovanni Cioffi, Nico Messikommer, and Davide Scaramuzza.
\newblock Reading in the dark with foveated event vision.
\newblock In \emph{IEEE/CVF Conference on Computer Vision and Pattern Recognition}, pages 5035--5043, 2025.

\bibitem[Brandli et~al.(2014)Brandli, Berner, Yang, Liu, and Delbruck]{brandli2014davis}
Christian Brandli, Raphael Berner, Minhao Yang, Shih-Chii Liu, and Tobi Delbruck.
\newblock A 240× 180 130 {dB} 3 µs latency global shutter spatiotemporal vision sensor.
\newblock \emph{IEEE Journal of Solid-State Circuits}, 49\penalty0 (10):\penalty0 2333--2341, 2014.

\bibitem[Cadena et~al.(2021)Cadena, Qian, Wang, and Yang]{cadena2021spade}
Pablo Rodrigo~Gantier Cadena, Yeqiang Qian, Chunxiang Wang, and Ming Yang.
\newblock Spade-{E2VID}: Spatially-adaptive denormalization for event-based video reconstruction.
\newblock \emph{IEEE Transactions on Image Processing}, 30:\penalty0 2488--2500, 2021.

\bibitem[Chaney et~al.(2023)Chaney, Cladera, Wang, Bisulco, Hsieh, Korpela, Kumar, Taylor, and Daniilidis]{chaney2023m3ed}
Kenneth Chaney, Fernando Cladera, Ziyun Wang, Anthony Bisulco, M.~Ani Hsieh, Christopher Korpela, Vijay Kumar, Camillo~J. Taylor, and Kostas Daniilidis.
\newblock {M3ED}: Multi-robot, multi-sensor, multi-environment event dataset.
\newblock In \emph{IEEE/CVF Conference on Computer Vision and Pattern Recognition Workshops}, pages 4016--4023, 2023.

\bibitem[{\c{C}}i{\c{c}}ek et~al.(2016){\c{C}}i{\c{c}}ek, Abdulkadir, Lienkamp, Brox, and Ronneberger]{cciccek20163d}
{\"O}zg{\"u}n {\c{C}}i{\c{c}}ek, Ahmed Abdulkadir, Soeren~S Lienkamp, Thomas Brox, and Olaf Ronneberger.
\newblock {3D} {U}-net: learning dense volumetric segmentation from sparse annotation.
\newblock In \emph{International Conference on Medical Image Computing and Computer-Assisted Intervention}, pages 424--432. Springer, 2016.

\bibitem[Dai et~al.(2022)Dai, Li, Zhou, Feng, and Loy]{dai2022flare7k}
Yuekun Dai, Chongyi Li, Shangchen Zhou, Ruicheng Feng, and Chen~Change Loy.
\newblock Flare7k: A phenomenological nighttime flare removal dataset.
\newblock In \emph{Advances in Neural Information Processing Systems}, pages 3926--3937, 2022.

\bibitem[Dai et~al.(2023)Dai, Luo, Zhou, Li, and Loy]{dai2023nighttime}
Yuekun Dai, Yihang Luo, Shangchen Zhou, Chongyi Li, and Chen~Change Loy.
\newblock Nighttime smartphone reflective flare removal using optical center symmetry prior.
\newblock In \emph{IEEE/CVF Conference on Computer Vision and Pattern Recognition}, pages 20783--20791, 2023.

\bibitem[Dai et~al.(2024)Dai, Li, Zhou, Feng, Luo, and Loy]{dai2024flare7k++}
Yuekun Dai, Chongyi Li, Shangchen Zhou, Ruicheng Feng, Yihang Luo, and Chen~Change Loy.
\newblock Flare7k++: Mixing synthetic and real datasets for nighttime flare removal and beyond.
\newblock \emph{IEEE Transactions on Pattern Analysis and Machine Intelligence}, 46\penalty0 (11):\penalty0 7041--7055, 2024.

\bibitem[Duan et~al.(2021{\natexlab{a}})Duan, Wang, Shi, Cossairt, Huang, and Katsaggelos]{duan2021guided}
Peiqi Duan, Zihao~W Wang, Boxin Shi, Oliver Cossairt, Tiejun Huang, and Aggelos~K Katsaggelos.
\newblock Guided event filtering: Synergy between intensity images and neuromorphic events for high performance imaging.
\newblock \emph{IEEE Transactions on Pattern Analysis and Machine Intelligence}, 44\penalty0 (11):\penalty0 8261--8275, 2021{\natexlab{a}}.

\bibitem[Duan et~al.(2021{\natexlab{b}})Duan, Wang, Zhou, Ma, and Shi]{duan2021eventzoom}
Peiqi Duan, Zihao~W Wang, Xinyu Zhou, Yi Ma, and Boxin Shi.
\newblock {EventZoom}: Learning to denoise and super resolve neuromorphic events.
\newblock In \emph{IEEE/CVF Conference on Computer Vision and Pattern Recognition}, pages 12824--12833, 2021{\natexlab{b}}.

\bibitem[Feng et~al.(2020)Feng, Lv, Liu, Zhang, Xiao, and Han]{feng2020event}
Yang Feng, Hengyi Lv, Hailong Liu, Yisa Zhang, Yuyao Xiao, and Chengshan Han.
\newblock Event density based denoising method for dynamic vision sensor.
\newblock \emph{Applied Sciences}, 10\penalty0 (6), 2020.

\bibitem[Finateu et~al.(2020)Finateu, Niwa, Matolin, Tsuchimoto, Mascheroni, Reynaud, Mostafalu, Brady, Chotard, LeGoff, et~al.]{finateu2020gen4}
Thomas Finateu, Atsumi Niwa, Daniel Matolin, Koya Tsuchimoto, Andrea Mascheroni, Etienne Reynaud, Pooria Mostafalu, Frederick Brady, Ludovic Chotard, Florian LeGoff, et~al.
\newblock 5.10 a 1280$\times$ 720 back-illuminated stacked temporal contrast event-based vision sensor with 4.86 $\mu$m pixels, 1.066 geps readout, programmable event-rate controller and compressive data-formatting pipeline.
\newblock In \emph{IEEE International Solid-State Circuits Conference}, pages 112--114, 2020.

\bibitem[Fowles(1989)]{fowles1989introduction}
Grant~R Fowles.
\newblock \emph{Introduction to modern optics}.
\newblock Courier Corporation, 1989.

\bibitem[Gallego et~al.(2020)Gallego, Delbr{\"u}ck, Orchard, Bartolozzi, Taba, Censi, Leutenegger, Davison, Conradt, Daniilidis, et~al.]{gallego2020event}
Guillermo Gallego, Tobi Delbr{\"u}ck, Garrick Orchard, Chiara Bartolozzi, Brian Taba, Andrea Censi, Stefan Leutenegger, Andrew~J Davison, J{\"o}rg Conradt, Kostas Daniilidis, et~al.
\newblock Event-based vision: A survey.
\newblock \emph{IEEE Transactions on Pattern Analysis and Machine Intelligence}, 44\penalty0 (1):\penalty0 154--180, 2020.

\bibitem[Geckeler et~al.(2025)Geckeler, Neugebauer, Muglikar, Scaramuzza, and Mintchev]{geckeler2025event}
Christian Geckeler, Niklas Neugebauer, Manasi Muglikar, Davide Scaramuzza, and Stefano Mintchev.
\newblock Event spectroscopy: Event-based multispectral and depth sensing using structured light.
\newblock \emph{arXiv preprint arXiv:2509.06741}, 2025.

\bibitem[Gehrig and Scaramuzza(2024)]{gehrig2024low}
Daniel Gehrig and Davide Scaramuzza.
\newblock Low-latency automotive vision with event cameras.
\newblock \emph{Nature}, 629\penalty0 (8014):\penalty0 1034--1040, 2024.

\bibitem[Gehrig et~al.(2020)Gehrig, Gehrig, Hidalgo-Carri{\'o}, and Scaramuzza]{gehrig2020video}
Daniel Gehrig, Mathias Gehrig, Javier Hidalgo-Carri{\'o}, and Davide Scaramuzza.
\newblock Video to events: Recycling video datasets for event cameras.
\newblock In \emph{IEEE/CVF Conference on Computer Vision and Pattern Recognition}, pages 3586--3595, 2020.

\bibitem[Gehrig et~al.(2021{\natexlab{a}})Gehrig, Aarents, Gehrig, and Scaramuzza]{gehrig2021dsec}
Mathias Gehrig, Willem Aarents, Daniel Gehrig, and Davide Scaramuzza.
\newblock {DSEC}: A stereo event camera dataset for driving scenarios.
\newblock \emph{IEEE Robotics and Automation Letters}, 6\penalty0 (3):\penalty0 4947--4954, 2021{\natexlab{a}}.

\bibitem[Gehrig et~al.(2021{\natexlab{b}})Gehrig, Millh{\"a}usler, Gehrig, and Scaramuzza]{gehrig2021raft}
Mathias Gehrig, Mario Millh{\"a}usler, Daniel Gehrig, and Davide Scaramuzza.
\newblock {E-RAFT}: Dense optical flow from event cameras.
\newblock In \emph{International Conference on 3D Vision}, pages 197--206, 2021{\natexlab{b}}.

\bibitem[Gehrig et~al.(2024)Gehrig, Muglikar, and Scaramuzza]{gehrig2024dense}
Mathias Gehrig, Manasi Muglikar, and Davide Scaramuzza.
\newblock Dense continuous-time optical flow from event cameras.
\newblock \emph{IEEE Transactions on Pattern Analysis and Machine Intelligence}, 46\penalty0 (7):\penalty0 4736--4746, 2024.

\bibitem[Goodman(2005)]{goodman2005introduction}
Joseph~W Goodman.
\newblock \emph{Introduction to Fourier optics}.
\newblock Roberts and Company Publishers, 2005.

\bibitem[Gu et~al.(2021)Gu, Li, Zhang, and Tian]{gu2021learn}
Daxin Gu, Jia Li, Yu Zhang, and Yonghong Tian.
\newblock How to learn a domain-adaptive event simulator?
\newblock In \emph{ACM International Conference on Multimedia}, pages 1275--1283, 2021.

\bibitem[Guo and Delbruck(2022)]{guo2022low}
Shasha Guo and Tobi Delbruck.
\newblock Low cost and latency event camera background activity denoising.
\newblock \emph{IEEE Transactions on Pattern Analysis and Machine Intelligence}, 45\penalty0 (1):\penalty0 785--795, 2022.

\bibitem[Hagenaars et~al.(2021)Hagenaars, Paredes-Vall{\'e}s, and De~Croon]{hagenaars2021self}
Jesse Hagenaars, Federico Paredes-Vall{\'e}s, and Guido De~Croon.
\newblock Self-supervised learning of event-based optical flow with spiking neural networks.
\newblock In \emph{Advances in Neural Information Processing Systems}, pages 7167--7179, 2021.

\bibitem[Hamaguchi et~al.(2023)Hamaguchi, Furukawa, Onishi, and Sakurada]{hamaguchi2023hmnet}
Ryuhei Hamaguchi, Yasutaka Furukawa, Masaki Onishi, and Ken Sakurada.
\newblock Hierarchical neural memory network for low latency event processing.
\newblock In \emph{IEEE/CVF Conference on Computer Vision and Pattern Recognition}, pages 22867--22876, 2023.

\bibitem[Han et~al.(2024{\natexlab{a}})Han, Li, Wei, and Ji]{han2024event}
Hanqian Han, Jianing Li, Henglu Wei, and Xiangyang Ji.
\newblock Event-{3DGS}: Event-based {3D} reconstruction using {3D} gaussian splatting.
\newblock \emph{Advances in Neural Information Processing Systems}, 37:\penalty0 128139--128159, 2024{\natexlab{a}}.

\bibitem[Han et~al.(2024{\natexlab{b}})Han, Lyu, Li, Wei, Li, Wei, Chen, and Ji]{han2024physical}
Haiqian Han, Jiacheng Lyu, Jianing Li, Henglu Wei, Cheng Li, Yajing Wei, Shu Chen, and Xiangyang Ji.
\newblock Physical-based event camera simulator.
\newblock In \emph{European Conference on Computer Vision}, pages 19--35. Springer, 2024{\natexlab{b}}.

\bibitem[Hidalgo-Carri{\'o} et~al.(2020)Hidalgo-Carri{\'o}, Gehrig, and Scaramuzza]{hidalgo2020learning}
Javier Hidalgo-Carri{\'o}, Daniel Gehrig, and Davide Scaramuzza.
\newblock Learning monocular dense depth from events.
\newblock In \emph{International Conference on 3D Vision}, pages 534--542, 2020.

\bibitem[Hu et~al.(2020)Hu, Binas, Neil, Liu, and Delbruck]{hu2020ddd20}
Yuhuang Hu, Jonathan Binas, Daniel Neil, Shih-Chii Liu, and Tobi Delbruck.
\newblock {DDD20} end-to-end event camera driving dataset: Fusing frames and events with deep learning for improved steering prediction.
\newblock In \emph{IEEE International Conference on Intelligent Transportation Systems}, pages 1--6, 2020.

\bibitem[Hu et~al.(2021)Hu, Liu, and Delbruck]{hu2021v2e}
Yuhuang Hu, Shih-Chii Liu, and Tobi Delbruck.
\newblock {V2E}: From video frames to realistic {DVS} events.
\newblock In \emph{IEEE/CVF Conference on Computer Vision and Pattern Recognition}, pages 1312--1321, 2021.

\bibitem[Hullin et~al.(2011)Hullin, Eisemann, Seidel, and Lee]{hullin2011physically}
Matthias Hullin, Elmar Eisemann, Hans-Peter Seidel, and Sungkil Lee.
\newblock Physically-based real-time lens flare rendering.
\newblock In \emph{ACM SIGGRAPH}, pages 1--10, 2011.

\bibitem[Im et~al.(2023)Im, Park, Kim, Son, Shin, and Lee]{im2023live}
Gyubeom Im, Keunjoo Park, Junseok Kim, Bongki Son, Seungchul Shin, and Haechang Lee.
\newblock Live demonstration: {PINK}: Polarity-based anti-flicker for event cameras.
\newblock In \emph{IEEE/CVF Conference on Computer Vision and Pattern Recognition Workshops}, pages 3901--3902, 2023.

\bibitem[Jiang et~al.(2024)Jiang, Chen, Pun, Wang, and Feng]{jiang2024mfdnet}
Yiguo Jiang, Xuhang Chen, Chi-Man Pun, Shuqiang Wang, and Wei Feng.
\newblock {MFDNet}: Multi-frequency deflare network for efficient nighttime flare removal.
\newblock \emph{The Visual Computer}, 40\penalty0 (11):\penalty0 7575--7588, 2024.

\bibitem[Jing et~al.(2024)Jing, Ding, Gao, Wang, Yan, Wang, Schaefer, Fang, Zhao, and Li]{jing2024hpl-ess}
Linglin Jing, Yiming Ding, Yunpeng Gao, Zhigang Wang, Xu Yan, Dong Wang, Gerald Schaefer, Hui Fang, Bin Zhao, and Xuelong Li.
\newblock {HPL-ESS}: Hybrid pseudo-labeling for unsupervised event-based semantic segmentation.
\newblock In \emph{IEEE/CVF Conference on Computer Vision and Pattern Recognition}, pages 23128--23137, 2024.

\bibitem[Joubert et~al.(2021)Joubert, Marcireau, Ralph, Jolley, Van~Schaik, and Cohen]{joubert2021event}
Damien Joubert, Alexandre Marcireau, Nic Ralph, Andrew Jolley, Andr{\'e} Van~Schaik, and Gregory Cohen.
\newblock Event camera simulator improvements via characterized parameters.
\newblock \emph{Frontiers in Neuroscience}, 15:\penalty0 702765, 2021.

\bibitem[Kim et~al.(2021)Kim, Bae, Park, Zhang, and Kim]{kim2021n-imagenet}
Junho Kim, Jaehyeok Bae, Gangin Park, Dongsu Zhang, and Young~Min Kim.
\newblock {N-ImageNet}: Towards robust, fine-grained object recognition with event cameras.
\newblock In \emph{IEEE/CVF International Conference on Computer Vision}, pages 2146--2156, 2021.

\bibitem[Kong et~al.(2024)Kong, Liu, Ng, Cottereau, and Ooi]{kong2024openess}
Lingdong Kong, Youquan Liu, Lai~Xing Ng, Benoit~R. Cottereau, and Wei~Tsang Ooi.
\newblock {OpenESS}: Event-based semantic scene understanding with open vocabularies.
\newblock In \emph{IEEE/CVF Conference on Computer Vision and Pattern Recognition}, pages 15686--15698, 2024.

\bibitem[Kong et~al.(2025{\natexlab{a}})Kong, Lu, Liang, Li, Dong, Hu, Ng, Ooi, and Cottereau]{kong2025talk2event}
Lingdong Kong, Dongyue Lu, Ao Liang, Rong Li, Yuhao Dong, Tianshuai Hu, Lai~Xing Ng, Wei~Tsang Ooi, and Benoit~R. Cottereau.
\newblock {Talk2Event}: Grounded understanding of dynamic scenes from event cameras.
\newblock In \emph{Advances in Neural Information Processing Systems}, 2025{\natexlab{a}}.

\bibitem[Kong et~al.(2025{\natexlab{b}})Kong, Lu, Liang, Li, Dong, Hu, Ng, Ooi, and Cottereau]{kong2025visual}
Lingdong Kong, Dongyue Lu, Ao Liang, Rong Li, Yuhao Dong, Tianshuai Hu, Lai~Xing Ng, Wei~Tsang Ooi, and Benoit~R. Cottereau.
\newblock Visual grounding from event cameras.
\newblock \emph{arXiv preprint arXiv:2509.09584}, 2025{\natexlab{b}}.

\bibitem[Kong et~al.(2025{\natexlab{c}})Kong, Lu, Xu, Ng, Ooi, and Cottereau]{kong2025eventfly}
Lingdong Kong, Dongyue Lu, Xiang Xu, Lai~Xing Ng, Wei~Tsang Ooi, and Benoit~R. Cottereau.
\newblock {EventFly}: Event camera perception from ground to the sky.
\newblock In \emph{IEEE/CVF Conference on Computer Vision and Pattern Recognition}, pages 1472--1484, 2025{\natexlab{c}}.

\bibitem[Li and Tian(2021)]{li2021recent}
Jianing Li and Yonghong Tian.
\newblock Recent advances in neuromorphic vision sensors: A survey.
\newblock \emph{Chinese Journal of Computers}, 44\penalty0 (6):\penalty0 1258--1286, 2021.

\bibitem[Li et~al.(2022)Li, Li, Zhu, Xiang, Huang, and Tian]{li2022asynchronous}
Jianing Li, Jia Li, Lin Zhu, Xijie Xiang, Tiejun Huang, and Yonghong Tian.
\newblock Asynchronous spatio-temporal memory network for continuous event-based object detection.
\newblock \emph{IEEE Transactions on Image Processing}, 31:\penalty0 2975--2987, 2022.

\bibitem[Li et~al.(2025)Li, Zhang, Han, and Ji]{li2025active}
Jianing Li, Yunjian Zhang, Haiqian Han, and Xiangyang Ji.
\newblock Active event-based stereo vision.
\newblock In \emph{IEEE/CVF Conference on Computer Vision and Pattern Recognition Conference}, pages 971--981, 2025.

\bibitem[Lin et~al.(2022)Lin, Ma, Guo, and Wen]{lin2022dvs}
Songnan Lin, Ye Ma, Zhenhua Guo, and Bihan Wen.
\newblock {DVS}-voltmeter: Stochastic process-based event simulator for dynamic vision sensors.
\newblock In \emph{European Conference on Computer Vision}, pages 578--593. Springer, 2022.

\bibitem[Lin et~al.(2023)Lin, Qiu, Shen, Zang, Liu, Bian, M{\"u}ller, Wang, et~al.]{lin2024e2pnet}
Xiuhong Lin, Changjie Qiu, Siqi Shen, Yu Zang, Weiquan Liu, Xuesheng Bian, Matthias M{\"u}ller, Cheng Wang, et~al.
\newblock {E2PNet}: Event to point cloud registration with spatio-temporal representation learning.
\newblock In \emph{Advances in Neural Information Processing Systems}, pages 18076--18089, 2023.

\bibitem[Lu et~al.(2025)Lu, Kong, Lee, Chane, and Ooi]{lu2025flexevent}
Dongyue Lu, Lingdong Kong, Gim~Hee Lee, Camille~Simon Chane, and Wei~Tsang Ooi.
\newblock {FlexEvent}: Towards flexible event-frame object detection at varying operational frequencies.
\newblock In \emph{Advances in Neural Information Processing Systems}, 2025.

\bibitem[Messikommer et~al.(2025)Messikommer, Fang, Gehrig, Cioffi, and Scaramuzza]{messikommer2025data}
Nico Messikommer, Carter Fang, Mathias Gehrig, Giovanni Cioffi, and Davide Scaramuzza.
\newblock Data-driven feature tracking for event cameras with and without frames.
\newblock \emph{IEEE Transactions on Pattern Analysis and Machine Intelligence}, 47\penalty0 (5):\penalty0 3706--3717, 2025.

\bibitem[Mildenhall et~al.(2021)Mildenhall, Srinivasan, Tancik, Barron, Ramamoorthi, and Ng]{mildenhall2021nerf}
Ben Mildenhall, Pratul~P Srinivasan, Matthew Tancik, Jonathan~T Barron, Ravi Ramamoorthi, and Ren Ng.
\newblock {NeRF}: Representing scenes as neural radiance fields for view synthesis.
\newblock \emph{Communications of the ACM}, 65\penalty0 (1):\penalty0 99--106, 2021.

\bibitem[Muglikar et~al.(2023)Muglikar, Bauersfeld, Moeys, and Scaramuzza]{muglikar2023event}
Manasi Muglikar, Leonard Bauersfeld, Diederik~Paul Moeys, and Davide Scaramuzza.
\newblock Event-based shape from polarization.
\newblock In \emph{IEEE/CVF Conference on Computer Vision and Pattern Recognition}, pages 1547--1556, 2023.

\bibitem[Muglikar et~al.(2025)Muglikar, Somasundaram, Dave, Charbon, Raskar, and Scaramuzza]{muglikar2025event}
Manasi Muglikar, Siddharth Somasundaram, Akshat Dave, Edoardo Charbon, Ramesh Raskar, and Davide Scaramuzza.
\newblock Event cameras meet spads for high-speed, low-bandwidth imaging.
\newblock \emph{IEEE Transactions on Pattern Analysis and Machine Intelligence}, 47\penalty0 (9):\penalty0 7886--7897, 2025.

\bibitem[Pantho et~al.(2022)Pantho, Mbongue, Bhowmik, and Bobda]{pantho2022event}
Md~Jubaer~Hossain Pantho, Joel~Mandebi Mbongue, Pankaj Bhowmik, and Christophe Bobda.
\newblock Event camera simulator design for modeling attention-based inference architectures.
\newblock \emph{Journal of Real-Time Image Processing}, 19\penalty0 (2):\penalty0 363--374, 2022.

\bibitem[Pellerito et~al.(2024)Pellerito, Cannici, Gehrig, Belhadj, Dubois-Matra, Casasco, and Scaramuzza]{pellerito2024deep}
Roberto Pellerito, Marco Cannici, Daniel Gehrig, Joris Belhadj, Olivier Dubois-Matra, Massimo Casasco, and Davide Scaramuzza.
\newblock Deep visual odometry with events and frames.
\newblock In \emph{IEEE/RSJ International Conference on Intelligent Robots and Systems}, pages 8966--8973, 2024.

\bibitem[Perot et~al.(2020)Perot, De~Tournemire, Nitti, Masci, and Sironi]{perot2020mpx1}
Etienne Perot, Pierre De~Tournemire, Davide Nitti, Jonathan Masci, and Amos Sironi.
\newblock Learning to detect objects with a 1 megapixel event camera.
\newblock In \emph{Advances in Neural Information Processing Systems}, pages 16639--16652, 2020.

\bibitem[Posch et~al.(2010)Posch, Matolin, and Wohlgenannt]{posch2010gen1}
Christoph Posch, Daniel Matolin, and Rainer Wohlgenannt.
\newblock A {QVGA} 143 {dB} dynamic range frame-free {PWM} image sensor with lossless pixel-level video compression and time-domain cds.
\newblock \emph{IEEE Journal of Solid-State Circuits}, 46\penalty0 (1):\penalty0 259--275, 2010.

\bibitem[Posch et~al.(2014)Posch, Serrano-Gotarredona, Linares-Barranco, and Delbruck]{posch2014retinomorphic}
Christoph Posch, Teresa Serrano-Gotarredona, Bernabe Linares-Barranco, and Tobi Delbruck.
\newblock Retinomorphic event-based vision sensors: Bioinspired cameras with spiking output.
\newblock \emph{Proceedings of the IEEE}, 102\penalty0 (10):\penalty0 1470--1484, 2014.

\bibitem[Qiao et~al.(2021)Qiao, Hancke, and Lau]{qiao2021light}
Xiaotian Qiao, Gerhard~P Hancke, and Rynson~WH Lau.
\newblock Light source guided single-image flare removal from unpaired data.
\newblock In \emph{IEEE/CVF International Conference on Computer Vision}, pages 4177--4185, 2021.

\bibitem[Rebecq et~al.(2018)Rebecq, Gehrig, and Scaramuzza]{rebecq2018esim}
Henri Rebecq, Daniel Gehrig, and Davide Scaramuzza.
\newblock {ESIM}: an open event camera simulator.
\newblock In \emph{Conference on Robot Learning}, pages 969--982, 2018.

\bibitem[Rios-Navarro et~al.(2023)Rios-Navarro, Guo, Gnaneswaran, Vijayakumar, Linares-Barranco, Aarrestad, Kastner, and Delbruck]{rios2023within}
Antonio Rios-Navarro, Shasha Guo, Abarajithan Gnaneswaran, Keerthivasan Vijayakumar, Alejandro Linares-Barranco, Thea Aarrestad, Ryan Kastner, and Tobi Delbruck.
\newblock Within-camera multilayer perceptron {DVS} denoising.
\newblock In \emph{IEEE/CVF Conference on Computer Vision and Pattern Recognition}, pages 3933--3942, 2023.

\bibitem[Rizzo et~al.(2022)Rizzo, Schuman, and Plank]{rizzo2022event}
Charles Rizzo, Catherine Schuman, and James Plank.
\newblock Event-based camera simulation wrapper for arcade learning environment.
\newblock In \emph{International Conference on Neuromorphic Systems}, pages 1--5, 2022.

\bibitem[Shi et~al.(2023)Shi, Li, Song, Wei, Zhang, Li, and Jin]{shi2023identifying}
Chenyang Shi, Yuzhen Li, Ningfang Song, Boyi Wei, Yibo Zhang, Wenzhuo Li, and Jing Jin.
\newblock Identifying light interference in event-based vision.
\newblock \emph{IEEE Transactions on Circuits and Systems for Video Technology}, 34\penalty0 (6):\penalty0 4800--4816, 2023.

\bibitem[Shi et~al.(2025)Shi, Wei, Wang, Liu, Zhang, Li, Song, and Jin]{shi2024polarity}
Chenyang Shi, Boyi Wei, Xiucheng Wang, Hanxiao Liu, Yibo Zhang, Wenzhuo Li, Ningfang Song, and Jing Jin.
\newblock Polarity-focused denoising for event cameras.
\newblock \emph{IEEE Transactions on Circuits and Systems for Video Technology}, 35\penalty0 (5):\penalty0 4370--4383, 2025.

\bibitem[Son et~al.(2017)Son, Suh, Kim, Jung, Kim, Shin, Park, Lee, Park, Woo, et~al.]{son2017gen3}
Bongki Son, Yunjae Suh, Sungho Kim, Heejae Jung, Jun-Seok Kim, Changwoo Shin, Keunju Park, Kyoobin Lee, Jinman Park, Jooyeon Woo, et~al.
\newblock 4.1 a 640$\times$ 480 dynamic vision sensor with a 9$\mu$m pixel and 300{Meps} address-event representation.
\newblock In \emph{IEEE International Solid-State Circuits Conference}, pages 66--67, 2017.

\bibitem[Stoffregen et~al.(2020)Stoffregen, Scheerlinck, Scaramuzza, Drummond, Barnes, Kleeman, and Mahony]{stoffregen2020reducing}
Timo Stoffregen, Cedric Scheerlinck, Davide Scaramuzza, Tom Drummond, Nick Barnes, Lindsay Kleeman, and Robert Mahony.
\newblock Reducing the sim-to-real gap for event cameras.
\newblock In \emph{European Conference on Computer Vision}, pages 534--549. Springer, 2020.

\bibitem[Sun et~al.(2025)Sun, Gehrig, Sakaridis, Gehrig, Liang, Sun, Xu, Wang, Van~Gool, and Scaramuzza]{sun2024unified}
Lei Sun, Daniel Gehrig, Christos Sakaridis, Mathias Gehrig, Jingyun Liang, Peng Sun, Zhijie Xu, Kaiwei Wang, Luc Van~Gool, and Davide Scaramuzza.
\newblock A unified framework for event-based frame interpolation with ad-hoc deblurring in the wild.
\newblock \emph{IEEE Transactions on Pattern Analysis and Machine Intelligence}, 47\penalty0 (4):\penalty0 2265--2279, 2025.

\bibitem[Sun et~al.(2022)Sun, Messikommer, Gehrig, and Scaramuzza]{sun2022ess}
Zhaoning Sun, Nico Messikommer, Daniel Gehrig, and Davide Scaramuzza.
\newblock {ESS}: Learning event-based semantic segmentation from still images.
\newblock In \emph{European Conference on Computer Vision}, pages 341--357, 2022.

\bibitem[Tulyakov et~al.(2022)Tulyakov, Bochicchio, Gehrig, Georgoulis, Li, and Scaramuzza]{tulyakov2022time}
Stepan Tulyakov, Alfredo Bochicchio, Daniel Gehrig, Stamatios Georgoulis, Yuanyou Li, and Davide Scaramuzza.
\newblock {Time Lens++}: Event-based frame interpolation with parametric non-linear flow and multi-scale fusion.
\newblock In \emph{IEEE/CVF Conference on Computer Vision and Pattern Recognition}, pages 17755--17764, 2022.

\bibitem[Valencia(2023)]{valencia2023optically}
Sebastian Valencia.
\newblock Optically enhancing event-based vision.
\newblock Master's thesis, The University of Arizona, 2023.

\bibitem[Vemprala et~al.(2021)Vemprala, Mian, and Kapoor]{vemprala2021representation}
Sai Vemprala, Sami Mian, and Ashish Kapoor.
\newblock Representation learning for event-based visuomotor policies.
\newblock In \emph{Advances in Neural Information Processing Systems}, pages 4712--4724, 2021.

\bibitem[Vitoria and Ballester(2019)]{vitoria2019automatic}
Patricia Vitoria and Coloma Ballester.
\newblock Automatic flare spot artifact detection and removal in photographs.
\newblock \emph{Journal of Mathematical Imaging and Vision}, 61\penalty0 (4):\penalty0 515--533, 2019.

\bibitem[Wang et~al.(2023)Wang, Li, Zhu, Zhang, Chen, Li, Wang, Tian, and Wu]{wang2023visevent}
Xiao Wang, Jianing Li, Lin Zhu, Zhipeng Zhang, Zhe Chen, Xin Li, Yaowei Wang, Yonghong Tian, and Feng Wu.
\newblock {VisEvent}: Reliable object tracking via collaboration of frame and event flows.
\newblock \emph{IEEE Transactions on Cybernetics}, 54\penalty0 (1):\penalty0 1997--2010, 2023.

\bibitem[Wang et~al.(2019)Wang, Du, Shen, Wu, Zhao, Sun, and Wen]{wang2019ev}
Yanxiang Wang, Bowen Du, Yiran Shen, Kai Wu, Guangrong Zhao, Jianguo Sun, and Hongkai Wen.
\newblock {EV-Gait}: Event-based robust gait recognition using dynamic vision sensors.
\newblock In \emph{IEEE/CVF Conference on Computer Vision and Pattern Recognition}, pages 6358--6367, 2019.

\bibitem[Wang et~al.(2022)Wang, Yuan, Ng, and Mahony]{wang2022linear}
Ziwei Wang, Dingran Yuan, Yonhon Ng, and Robert Mahony.
\newblock A linear comb filter for event flicker removal.
\newblock In \emph{IEEE International Conference on Robotics and Automation}, pages 398--404, 2022.

\bibitem[Wang et~al.(2020)Wang, Duan, Cossairt, Katsaggelos, Huang, and Shi]{wang2020joint}
Zihao~W Wang, Peiqi Duan, Oliver Cossairt, Aggelos Katsaggelos, Tiejun Huang, and Boxin Shi.
\newblock Joint filtering of intensity images and neuromorphic events for high-resolution noise-robust imaging.
\newblock In \emph{IEEE/CVF Conference on Computer Vision and Pattern Recognition}, pages 1609--1619, 2020.

\bibitem[Wolny et~al.(2020)Wolny, Cerrone, Vijayan, Tofanelli, Barro, Louveaux, Wenzl, Strauss, Wilson-Sánchez, Lymbouridou, Steigleder, Pape, Bailoni, Duran-Nebreda, Bassel, Lohmann, Tsiantis, Hamprecht, Schneitz, Maizel, and Kreshuk]{10.7554/eLife.57613}
Adrian Wolny, Lorenzo Cerrone, Athul Vijayan, Rachele Tofanelli, Amaya~Vilches Barro, Marion Louveaux, Christian Wenzl, Sören Strauss, David Wilson-Sánchez, Rena Lymbouridou, Susanne~S Steigleder, Constantin Pape, Alberto Bailoni, Salva Duran-Nebreda, George~W Bassel, Jan~U Lohmann, Miltos Tsiantis, Fred~A Hamprecht, Kay Schneitz, Alexis Maizel, and Anna Kreshuk.
\newblock Accurate and versatile {3D} segmentation of plant tissues at cellular resolution.
\newblock \emph{eLife}, 9:\penalty0 e57613, 2020.

\bibitem[Wu et~al.(2020)Wu, Ma, Li, Dong, and Shi]{wu2020probabilistic}
Jinjian Wu, Chuanwei Ma, Leida Li, Weisheng Dong, and Guangming Shi.
\newblock Probabilistic undirected graph based denoising method for dynamic vision sensor.
\newblock \emph{IEEE Transactions on Multimedia}, 23:\penalty0 1148--1159, 2020.

\bibitem[Wu et~al.(2021)Wu, He, Xue, Garg, Chen, Veeraraghavan, and Barron]{wu2021train}
Yicheng Wu, Qiurui He, Tianfan Xue, Rahul Garg, Jiawen Chen, Ashok Veeraraghavan, and Jonathan~T Barron.
\newblock How to train neural networks for flare removal.
\newblock In \emph{IEEE/CVF International Conference on Computer Vision}, pages 2239--2247, 2021.

\bibitem[Zhang et~al.(2023{\natexlab{a}})Zhang, Ge, Song, and Lam]{zhang2023neuromorphic}
Pei Zhang, Zhou Ge, Li Song, and Edmund~Y Lam.
\newblock Neuromorphic imaging with density-based spatiotemporal denoising.
\newblock \emph{IEEE Transactions on Computational Imaging}, 9:\penalty0 530--541, 2023{\natexlab{a}}.

\bibitem[Zhang et~al.(2023{\natexlab{b}})Zhang, Cui, Chai, Yu, Dasgupta, Mahbub, and Rahman]{zhang2023v2ce}
Zhongyang Zhang, Shuyang Cui, Kaidong Chai, Haowen Yu, Subhasis Dasgupta, Upal Mahbub, and Tauhidur Rahman.
\newblock {V2CE}: Video to continuous events simulator.
\newblock \emph{arXiv preprint arXiv:2309.08891}, 2023{\natexlab{b}}.

\bibitem[Zheng et~al.(2023)Zheng, Liu, Lu, Hua, Pan, Zhang, Tao, and Wang]{zheng2023deep}
Xu Zheng, Yexin Liu, Yunfan Lu, Tongyan Hua, Tianbo Pan, Weiming Zhang, Dacheng Tao, and Lin Wang.
\newblock Deep learning for event-based vision: A comprehensive survey and benchmarks.
\newblock \emph{arXiv preprint arXiv:2302.08890}, 2023.

\bibitem[Zhou et~al.(2023)Zhou, Liang, Chen, Huang, Yang, and Li]{zhou2023improving}
Yuyan Zhou, Dong Liang, Songcan Chen, Sheng-Jun Huang, Shuo Yang, and Chongyi Li.
\newblock Improving lens flare removal with general-purpose pipeline and multiple light sources recovery.
\newblock In \emph{IEEE/CVF International Conference on Computer Vision}, pages 12969--12979, 2023.

\bibitem[Zhu et~al.(2018)Zhu, Thakur, {\"O}zaslan, Pfrommer, Kumar, and Daniilidis]{zhu2018multivehicle}
Alex~Zihao Zhu, Dinesh Thakur, Tolga {\"O}zaslan, Bernd Pfrommer, Vijay Kumar, and Kostas Daniilidis.
\newblock The multivehicle stereo event camera dataset: An event camera dataset for {3D} perception.
\newblock \emph{IEEE Robotics and Automation Letters}, 3\penalty0 (3):\penalty0 2032--2039, 2018.

\bibitem[Zhu et~al.(2021)Zhu, Wang, Khant, and Daniilidis]{zhu2021eventgan}
Alex~Zihao Zhu, Ziyun Wang, Kaung Khant, and Kostas Daniilidis.
\newblock {EventGAN}: Leveraging large-scale image datasets for event cameras.
\newblock In \emph{IEEE International Conference on Computational Photography}, pages 1--11, 2021.

\bibitem[Zubić et~al.(2023)Zubić, Gehrig, Gehrig, and Scaramuzza]{zubic2023from}
Nikola Zubić, Daniel Gehrig, Mathias Gehrig, and Davide Scaramuzza.
\newblock From chaos comes order: Ordering event representations for object recognition and detection.
\newblock In \emph{IEEE/CVF International Conference on Computer Vision}, pages 12846--128567, 2023.

\end{thebibliography}
}

% Appendix
\clearpage
\onecolumn
\setcounter{section}{0}
\renewcommand{\thesection}{\Alph{section}}

% Table of Contents
\section*{Table of Contents}
\startcontents[appendices]
\printcontents[appendices]{l}{1}{\setcounter{tocdepth}{3}}

\clearpage\clearpage
\section{~The E-Deflare Benchmark}
\label{app:data_acquisition}

In this section, we provide an in-depth exposition of the data generation methodologies and physical acquisition protocols established for the \textbf{E-Deflare Benchmark}. This benchmark was meticulously designed to address the critical scarcity of high-quality, paired data required for training and evaluating event-based flare removal algorithms.

\vspace{0.1cm}
\subsection{~Benchmark Overview}
The \textbf{E-Deflare Benchmark} constitutes a holistic suite of datasets tailored to facilitate different stages of model development and rigorous evaluation. It comprises three complementary components: the large-scale simulated training set (\textbf{E-Flare-2.7K}) for robust feature learning; the first-of-its-kind paired real-world test set (\textbf{E-Flare-R}) for sim-to-real validation; and a curated collection of qualitative real-world sequences (\textbf{DSEC-Flare}) from the popular DSEC dataset for assessing generalization in complex, in-the-wild scenarios.

\vspace{0.1cm}
\subsection{~E-Flare-2.7K: A Simulated Dataset}
\label{app:eflare_details}

The \textbf{E-Deflare-2.7K} dataset constitutes the foundational pillar for training our restoration network. It is constructed using our novel \textbf{physics-driven simulator}, a pipeline specifically engineered to synthesize physically grounded paired data. The core mechanism involves fusing dynamic, simulated flare events with authentic real-world background events via the \emph{Probabilistic Non-Linear Event Suppression (PNL-ES)} operator. This approach ensures that the interaction between signal and artifact mimics physical reality rather than simple linear addition.

\begin{figure}[b]
    \centering
    \vspace{0.1cm}
    \includegraphics[width=\textwidth]{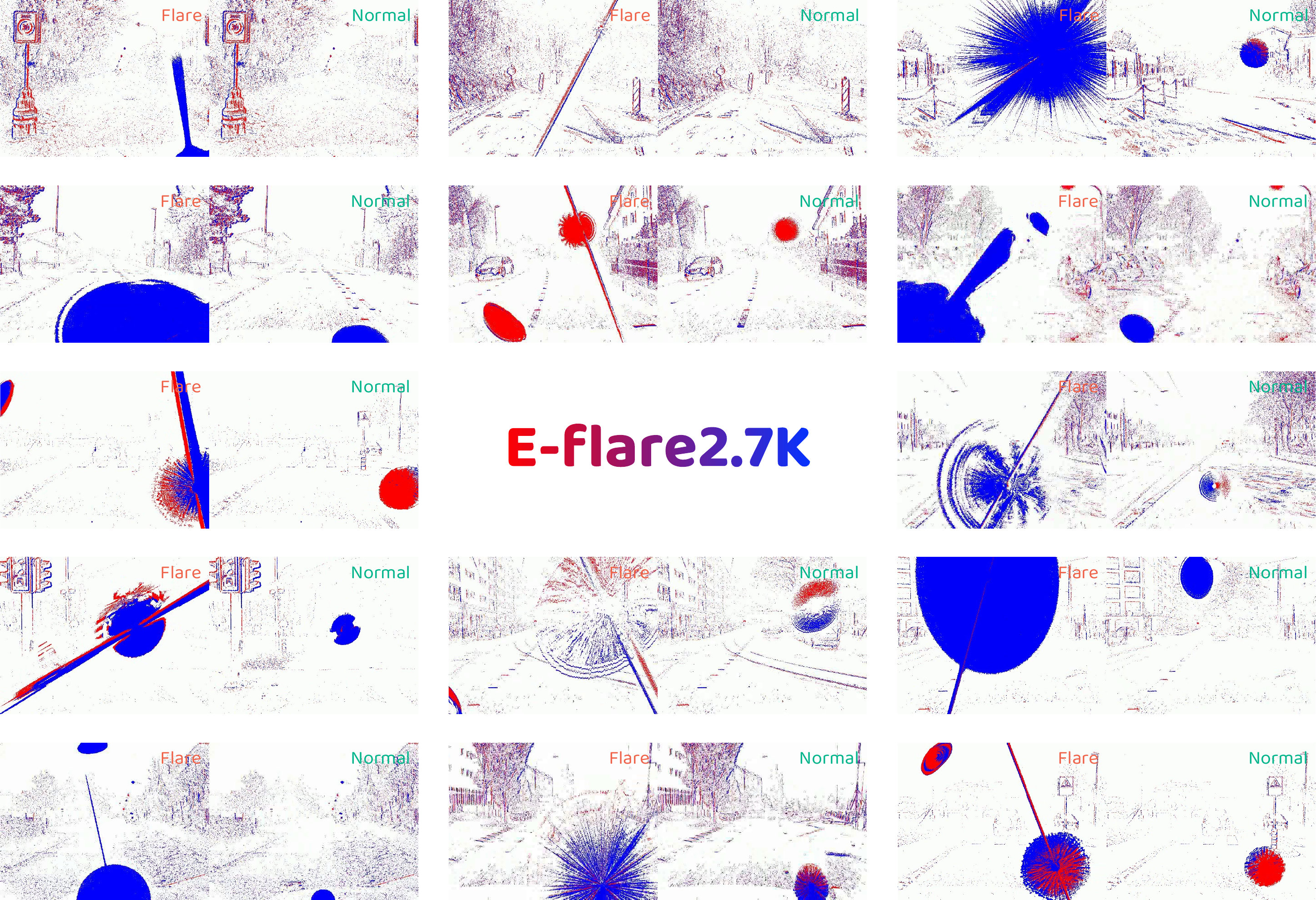}
    \vspace{-0.55cm}
    \caption{\textbf{Examples from E-Flare2.7K}. This is a large-scale paired dataset that combines synthetic event data under \textbf{\textcolor{ev_red}{flare}} (left) and \textbf{\textcolor{ev_green}{normal}} (right) conditions, respectively. Best viewed in colors and zoomed-in for details.}
    \label{fig:dataset_e-flare}
\end{figure}

\vspace{0.1cm}
\noindent\textbf{Background Event Scenes.}
To guarantee that our synthetic data faithfully reflects the complexity of real-world environments, the background events ($\mathbf{E}_{\mathrm{bg}}$) are strictly sampled from the DSEC dataset~\cite{gehrig2021dsec}. We utilize seven distinct, long-duration sequences that capture diverse urban driving environments (\eg, \texttt{zurich\_city\_00\_a}, \texttt{interlaken\_00\_c}). For each training sample, we extract a random temporal window ranging from $50$ to $100$~ms. This randomized extraction strategy provides a rich variety of background motion patterns and structural densities, thereby preventing the model from overfitting to specific scene geometries or static background configurations.

\vspace{0.1cm}
\noindent\textbf{Flare Event Generation.}
The synthesis of the paired flare events ($\mathbf{E}_{\textcolor{ev_red}{\mathrm{\mathbf{f}}}}$) and the corresponding clean light source events ($\mathbf{E}_{\mathrm{light}}$) is strictly governed by a deterministic ``script-based'' synchronization strategy. This architectural choice is paramount for ensuring perfect spatio-temporal alignment between the input and ground truth. 

The generation process involves several key steps:

\begin{itemize}
    \item \textbf{Source Assets:} We utilize high-quality paired static assets sourced from Flare7K++~\cite{dai2024flare7k++}. These assets cover a wide spectrum of optical artifacts, including both scattering and reflective flare types, alongside their corresponding clean light source templates, providing a strong visual basis for simulation.

    \vspace{0.1cm}
    \item \textbf{Deterministic Synchronization:} To maintain absolute consistency, a global parameter script is generated for each sample. This script defines the precise motion trajectory, intensity flicker profile, and geometric transformations to be applied. Both $\mathbf{E}_{\textcolor{ev_red}{\mathrm{\mathbf{f}}}}$ and $\mathbf{E}_{\mathrm{light}}$ are rendered by executing this identical script, guaranteeing that the flare artifacts and the primary light source share the exact same temporal evolution and spatial positioning, effectively eliminating alignment errors.

    \vspace{0.1cm}
    \item \textbf{Dynamic Simulation:} To bridge the domain gap between static source images and dynamic event streams, we introduce multiple layers of randomization to simulate realistic physical variations:
    \begin{itemize}
        \item \textit{Motion Profiles:} To approximate the complex kinetics of real-world driving, we simulate realistic camera ego-motion by applying smooth linear trajectories. These are generated with random start positions, arbitrary directions ($0$--$360$°), and variable travel distances (up to $180$ pixels).
        \item \textit{Geometric Augmentations:} To enhance morphological diversity and robustness to scale/rotation changes, we apply random rotations ($0$--$360$°), scaling ($0.8$--$1.5$$\times$), translations ($\pm$$20$\%), and shearing ($\pm$$20$°).
        \item \textit{Intensity Flicker:} To emulate the characteristics of artificial lighting in urban nights, we apply periodic flickering. The base frequencies are sampled from $100$--$140$~Hz with varied waveforms (sine, square, triangle). Additionally, stable light sources are simulated by disabling flicker in $30\%$ of the samples to ensure coverage of non-flickering scenarios.
    \end{itemize}

    \vspace{0.1cm}
    \item \textbf{Hybrid Flare Rendering:} The final video generation employs a hybrid approach to maximize visual fidelity. The \textbf{scattering flare} is derived by animating the Flare7K++ assets, while the complex \textbf{reflective flare} is synthesized via a high-performance 2D shader. This shader is applied in $90\%$ of sequences to accurately model complex, spatially variant caustic patterns that are difficult to capture with static images alone.

    \vspace{0.1cm}
    \item \textbf{Event Stream Generation:} The rendered high-framerate videos (up to $\mathbf{6000}$ \textbf{FPS}) are converted into asynchronous event streams using the physics-based DVS-Voltmeter simulator~\cite{lin2022dvs}. During this process, we randomize the contrast threshold parameters to vary the resulting event density, further improving the network's robustness to varying sensor noise levels and sensitivity settings.
\end{itemize}

\vspace{0.1cm}
As visually depicted in Fig.~\ref{fig:dataset_e-flare}, the resulting dataset contains $\mathbf{2{,}720}$ \textbf{paired} $\mathbf{20}$~\textbf{ms} \textbf{samples} with a spatial resolution of $640 \times 480$. These are systematically split into $\mathbf{2{,}545}$ samples \textbf{for training} and $\mathbf{175}$ samples \textbf{for testing}.

\begin{figure}[t]
    \centering
    \includegraphics[width=\textwidth]{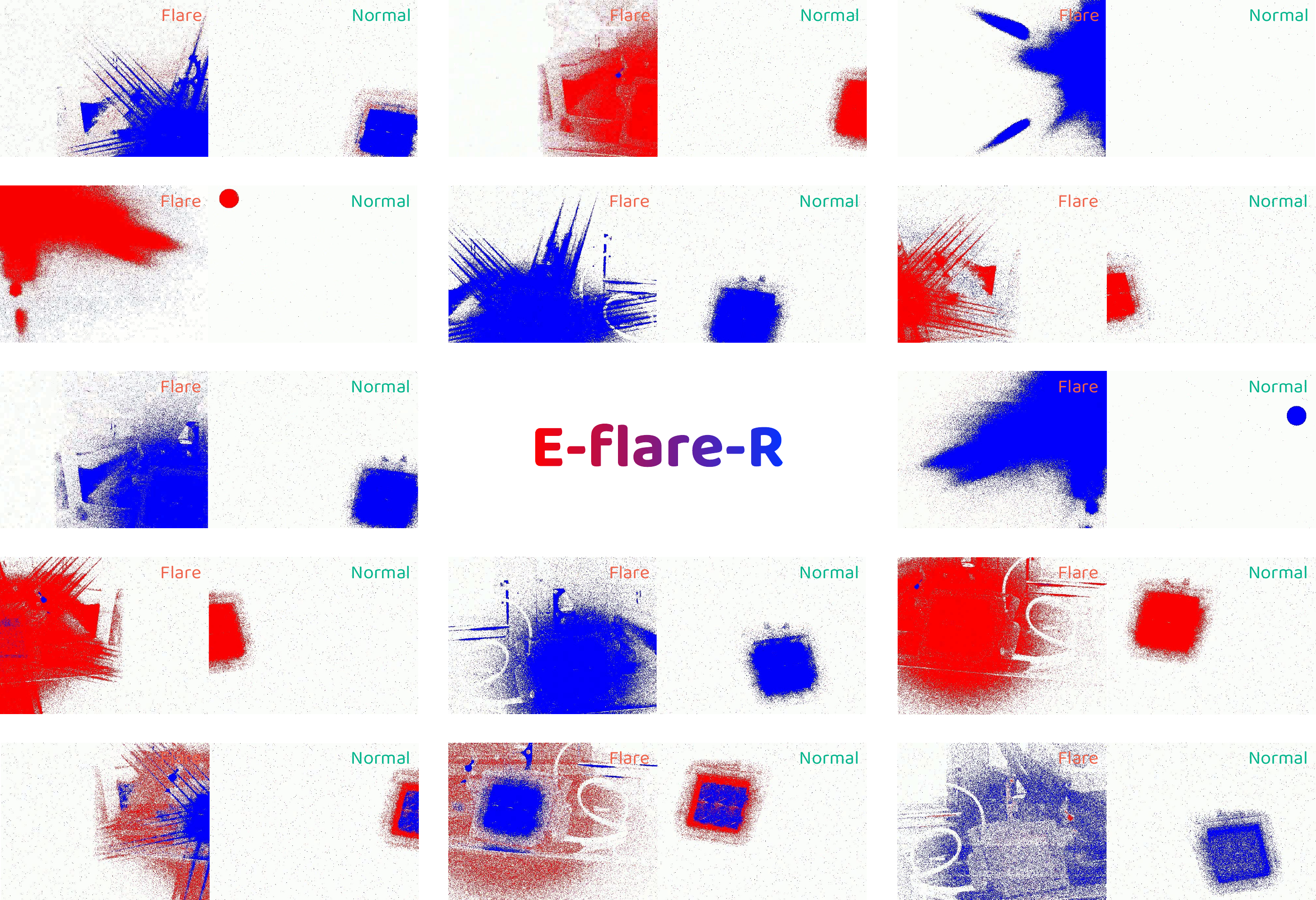}
    \vspace{-0.55cm}
    \caption{\textbf{Examples from E-Flare-R}. This is a real-world paired dataset that combines the acquired event data under \textbf{\textcolor{ev_red}{flare}} (left) and \textbf{\textcolor{ev_green}{normal}} (right) conditions, respectively. Best viewed in colors and zoomed-in for details.}
    \label{fig:dataset_e-flare-r}
\end{figure}

\begin{figure}[t]
    \centering
    \includegraphics[width=\textwidth]{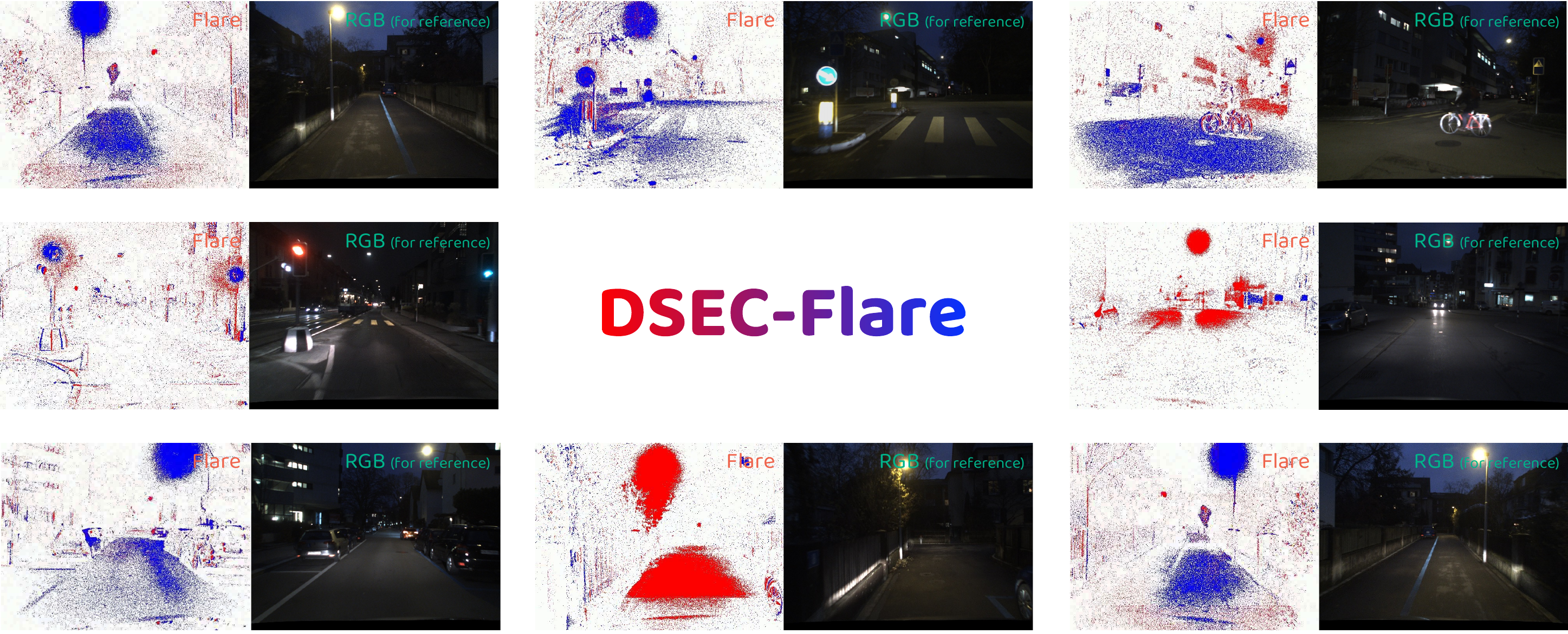}
    \vspace{-0.55cm}
    \caption{\textbf{Examples from DSEC-Flare}. This is a real-world dataset that showcases typical lens flares induced by various light sources. For each example, we show the raw event stream and a corresponding RGB image to illustrate the light sources causing the artifacts. Best viewed in colors and zoomed-in for details.}
    \label{fig:dataset_dsec-flare}
\end{figure}

\vspace{0.1cm}
\subsection{~E-Flare-R: A Real-World Paired Dataset}
\label{app:eflare_r_details}

The \textbf{E-Flare-R} dataset stands as a pivotal contribution to the field, representing the \textbf{first paired real-world benchmark} specifically designed for the event camera de-flaring task. It directly addresses the challenge of objective evaluation by capturing the exact same scene radiance under distinct optical impulse responses.

\vspace{0.1cm}
\noindent\textbf{Acquisition Setup.}
Data acquisition is performed using a Prophesee EVK4-HD event camera ($1280 \times 720$). To systematically introduce controllable and repeatable flare artifacts into the scene, we engineered a custom rig equipped with a set of removable optical filters, specifically including a six-point star filter and various random-pattern filters. For illumination, we positioned high-intensity light sources within the Field-of-View (FoV). These sources include a frequency-controllable point light and a square Tungsten lamp, allowing us to generate diverse flare artifacts under controlled experimental conditions.

\vspace{0.1cm}
\noindent\textbf{Capture Protocol.}
To obtain high-fidelity paired data, we execute a strict two-pass recording protocol for each scene:
\begin{enumerate}
    \item \textbf{Observation Pass ($\mathbf{E}_{\mathrm{ob}}$):}\\In the first pass, the scene is recorded with the optical filter mounted (or the strobe light activated). This pass captures the scene corrupted by authentic lens flare artifacts, serving as the input for the network.

    \vspace{0.1cm}
    \item \textbf{Reference Pass ($\tilde{\mathbf{E}}_{\mathrm{gt}}$):}\\Immediately following the observation pass, the filter is removed (or the strobe deactivated), and the scene is recorded again for the same duration. The camera remains rigidly static during this process to minimize spatial misalignment between the pairs.
\end{enumerate}

\vspace{0.1cm}
\noindent\textbf{Post-Processing Pipeline.}
The raw recordings undergo a rigorous, multi-stage processing pipeline to yield the final spatially and temporally aligned pairs suitable for benchmarking.
\begin{itemize}
    \item \textbf{Metric-Guided Temporal Alignment:} Since hardware triggers are unavailable for the two separate passes, we employ a signal-based software alignment method. We exploit the high-frequency signature of the strobe light, which creates a distinct, periodic pattern in the event polarity ratio. By maximizing the cross-correlation of this polarity signature between the two streams, we achieve precise, sub-millisecond temporal synchronization.

    \vspace{0.1cm}
    \item \textbf{Reference Cleaning:} Although the reference stream $\tilde{\mathbf{E}}_{\mathrm{gt}}$ is captured without the filter, it may still contain minor artifacts or stray noise. It serves as the base for the ground truth. To ensure purity, we apply a spatial mask derived from the known light source geometry to remove any residual glare events, ensuring that $\mathbf{E}_{\mathrm{gt}}$ represents only the clean scene structure.

    \vspace{0.1cm}
    \item \textbf{Noise Injection:} To prevent the network from overfitting to the artificially clean background of the masked ground truth, we inject randomized spatio-temporal noise. This noise is calibrated to match the statistical distribution of the sensor's intrinsic background activity.

    \vspace{0.1cm}
    \item \textbf{Sampling:} Finally, the aligned streams are center-cropped to $640 \times 480$ to remove lens edge effects and sliced into $100$~ms windows. This rigorous process results in $30$ high-quality paired sequences for robust evaluation.
\end{itemize}

\vspace{0.1cm}
Fig.~\ref{fig:dataset_e-flare-r} presents representative samples from \textbf{E-Flare-R}, showcasing the diversity of artifacts captured under varying combinations of light sources and optical filters.

\vspace{0.1cm}
\subsection{~DSEC-Flare: Curated Real-World Examples}
\label{app:dsec_flare_details}

To further validate the practical utility of our method in unconstrained environments, we introduce \textbf{DSEC-Flare}, a curated subset of the widely used DSEC dataset~\cite{gehrig2021dsec}. These sequences were specifically selected to highlight the prevalence and severity of lens flare in standard autonomous driving datasets.

\vspace{0.1cm}
It consists of challenging sequences extracted from the left camera ($640 \times 480$), which are categorized into two primary groups based on the nature of the flare source:

\vspace{0.1cm}
\begin{itemize}
    \item \textbf{Periodic Streetlight Flare:}\\This category encompasses sequences such as \texttt{zurich\_city\_03\_a}, \texttt{zurich\_city\_10\_b}, and \texttt{zurich\_city\_12\_a}. These scenarios are dominated by high-intensity streetlights that create strong, periodic flare bursts as the camera passes underneath them. The primary challenge in these sequences is the intense strobing effect caused by the rapid relative motion and proximity to these powerful, stationary light sources.

    \vspace{0.1cm}
    \item \textbf{Complex Urban Flare:}\\This group, comprising sequences like \texttt{zurich\_city\_09\_c} and \texttt{zurich\_city\_09\_d}, represents the complex lighting environments typical of dense city driving. The artifacts here arise from a diverse mixture of light sources, including oncoming vehicle headlights, traffic signals, and storefront illumination. These sequences are crucial for testing the model's ability to handle multiple, overlapping flare patterns with varying intensities and temporal dynamics.
\end{itemize}

\vspace{0.1cm}
Fig.~\ref{fig:dataset_dsec-flare} provides typical examples from \textbf{DSEC-Flare}. It contains representative DSEC sequences featuring flare artifacts caused by diverse types of light sources.
\clearpage\clearpage
\section{~Detailed Theoretical Derivation}
\label{app:derivation}

This section provides a step-by-step derivation of the theoretical forward model presented in the main text. We elaborate on the mathematical formulation to offer a deeper understanding of the non-linear suppression mechanism governing flare corruption in event streams.

\vspace{0.1cm}
\subsection{~Physical Origin of Lens Flare in Event Cameras}

To understand why event cameras are susceptible to lens flare, we first formalize the imaging process. The output of any camera is the result of a lens system, $\mathcal{L}$, forming an irradiance map on a sensor, $\mathcal{S}$. An ideal lens, $\mathcal{L}_{\mathrm{ideal}}$, would perfectly map scene radiance to a clean sensor irradiance, $\mathbf{I}_{\mathrm{gt}}(x, y, t)$. However, a real lens, $\mathcal{L}_{\mathrm{real}}$, introduces optical artifacts, \ie:
\begin{equation}
    \mathbf{I}_{\mathrm{ob}}(x, y, t) = \mathcal{L}_{\mathrm{real}}(\mathbf{I}_{\mathrm{gt}}(x, y, t))~,
\end{equation}
where $\mathbf{I}_{\mathrm{ob}}$ is the final, observed irradiance. Lens flare is this corruption. As flare is a high-intensity phenomenon, it can be effectively approximated by a linear superposition in the intensity domain~\cite{dai2022flare7k, wu2021train, dai2024flare7k++}:
\begin{equation}
    \mathbf{I}_{\mathrm{ob}}(x, y, t) \approx \mathbf{I}_{\mathrm{bg}}(x, y, t) + \mathbf{I}_{\textcolor{ev_red}{\mathrm{\mathbf{f}}}}(x, y, t)~, 
    \label{app:eq:image_superposition_final}
\end{equation}
where the scene is decomposed into the background $\mathbf{I}_{\mathrm{bg}}$ and the flare pattern $\mathbf{I}_{\textcolor{ev_red}{\mathrm{\mathbf{f}}}}$.

This corruption occurs in the lens, \emph{before} the sensor. The sensor, $\mathcal{S}$, determines the camera type. An event camera sensor, $\mathcal{S}_{\mathrm{evt}}$, operates on this corrupted irradiance, outputting an asynchronous stream of events, $\mathbf{E}_{\mathrm{ob}}(t) = \mathcal{S}_{\mathrm{evt}}(\mathbf{I}_{\mathrm{ob}}(t))$. This stream is represented as a series of signed impulses:
\begin{equation}
    \mathbf{E}(t) = \sum_{i} p_i \delta(t - t_i)~,
    \label{app:eq:event_stream_definition}
\end{equation}
where each event consists of a timestamp $t_i$ and polarity $p_i \in \{-1, 1\}$.

The operator $\mathcal{S}_{\mathrm{evt}}$ is a stateful process where each pixel tracks the log-intensity $L(t) = \log(\mathbf{I}(t))$ and fires an event when the change since the last event crosses a contrast threshold $c$:
\begin{equation}
    |L(t_i) - L(t_{\mathrm{last}})| \ge c~.
\end{equation}
The ideal output would be $\mathbf{E}_{\mathrm{gt}} = \mathcal{S}_{\mathrm{evt}}(\mathbf{I}_{\mathrm{gt}}(t))$. However, the sensor receives the corrupted signal, so the actual output is $\mathbf{E}_{\mathrm{ob}} = \mathcal{S}_{\mathrm{evt}}(\mathbf{I}_{\mathrm{ob}}(t))$. Since the flare intensity $\mathbf{I}_{\textcolor{ev_red}{\mathrm{\mathbf{f}}}}(t)$ is a dynamic signal, it generates its own spurious events, corrupting the dynamics of the real scene. Thus, while its differential mechanism avoids saturation, \textbf{an event camera does not confer immunity to lens flare}.

This establishes event de-flaring as an inverse problem: recovering the clean stream $\mathbf{E}_{\mathrm{gt}}$ from the observation $\mathbf{E}_{\mathrm{ob}}$. The complexity of this non-linear, spatio-temporal mapping motivates a learning-based approach, which in turn requires a forward model for data synthesis.

\vspace{0.1cm}
\subsection{~A Theoretical Model of Flare in the Event Domain}
\label{app:ssec:event_synthesis_model}

To derive how the intensity superposition (Eq.~\ref{app:eq:image_superposition_final}) translates into the event domain, we use the integral form of the event camera model:
\begin{equation}
    L(t) = c \int_{0}^{t} \mathbf{E}(\tau) \, d\tau + L(0) + \varepsilon(t)~,
    \label{app:eq:sensor_model_integral_redef}
\end{equation}
where $L(0)$ is the initial log-intensity and $\varepsilon(t)$ is a bounded quantization error ($|\varepsilon(t)| < c$). The time derivative of this relation is key:
\begin{equation}
    \frac{d}{dt} L(t) = c \cdot \mathbf{E}(t) + \frac{d}{dt} \varepsilon(t)~.
    \label{app:eq:sensor_model_derivative_redef}
\end{equation}
This decomposes the rate of log-intensity change into an impulsive component (the event stream) and a continuous, sub-threshold component. We now derive the model for the observed event stream, $\mathbf{E}_{\mathrm{ob}}$. Starting from the linear intensity superposition and converting to the log domain gives:
\begin{equation}
    L_{\mathrm{ob}}(t) = \log(\mathbf{I}_{\mathrm{bg}}(t) + \mathbf{I}_{\textcolor{ev_red}{\mathrm{\mathbf{f}}}}(t)) = \log(e^{L_{\mathrm{bg}}(t)} + e^{L_{\textcolor{ev_red}{\mathrm{\mathbf{f}}}}(t)})~.
\end{equation}
Applying the chain rule to the time derivative of $L_{\mathrm{ob}}(t)$ yields:
\begin{equation}
    \frac{d}{dt} L_{\mathrm{ob}}(t) = w_{\mathrm{bg}}(t)\frac{d}{dt}L_{\mathrm{bg}}(t) + w_{\textcolor{ev_red}{\mathrm{\mathbf{f}}}}(t)\frac{d}{dt}L_{\textcolor{ev_red}{\mathrm{\mathbf{f}}}}(t)~,
\end{equation}
where the dynamic weights $w(t)$ are determined by the instantaneous ratio of their respective linear intensities, identical to the main text:
\begin{equation}
    w_{\mathrm{bg}}(t) = \frac{\mathbf{I}_{\mathrm{bg}}(t)}{\mathbf{I}_{\mathrm{bg}}(t) + \mathbf{I}_{\textcolor{ev_red}{\mathrm{\mathbf{f}}}}(t)}, \quad w_{\textcolor{ev_red}{\mathrm{\mathbf{f}}}}(t) = 1 - w_{\mathrm{bg}}(t)~.
\end{equation}
Substituting the sensor model from Eq.~\ref{app:eq:sensor_model_derivative_redef} for each component, we arrive at the complete expression for the observed log-intensity change, \ie:
\begin{equation}
\begin{split}
    c \cdot \mathbf{E}_{\mathrm{ob}}(t) + \frac{d}{dt}\varepsilon_{\mathrm{ob}}(t) = w_{\mathrm{bg}}(t)\left[c \cdot \mathbf{E}_{\mathrm{bg}}(t) + \frac{d}{dt}\varepsilon_{\mathrm{bg}}(t)\right] 
    + w_{\textcolor{ev_red}{\mathrm{\mathbf{f}}}}(t)\left[c \cdot \mathbf{E}_{\textcolor{ev_red}{\mathrm{\mathbf{f}}}}(t) + \frac{d}{dt}\varepsilon_{\textcolor{ev_red}{\mathrm{\mathbf{f}}}}(t)\right]~.
\end{split}
\label{app:eq:full_event_mixing_redef}
\end{equation}
This physically grounded equation reveals our key insight: the observed event stream results from a \textbf{non-linear fusion} of the source streams, dynamically weighted by their relative intensities.

In an idealized scenario where sub-threshold drifts are omitted ($\frac{d}{dt}\varepsilon(t) \approx 0$), the core interaction simplifies to:
\begin{equation}
    \mathbf{E}_{\mathrm{ideal}}(t) = w_{\mathrm{bg}}(t)\mathbf{E}_{\mathrm{bg}}(t) + w_{\textcolor{ev_red}{\mathrm{\mathbf{f}}}}(t)\mathbf{E}_{\textcolor{ev_red}{\mathrm{\mathbf{f}}}}(t)~.
    \label{app:eq:hypo_stream_final_redef}
\end{equation}
This expression mathematically formalizes the \textbf{non-linear suppression effect} described in the main paper. However, this idealized stream $\mathbf{E}_{\mathrm{ideal}}(t)$ is a mathematical abstraction with non-standard amplitudes.

To obtain the final, physically realizable output, the event generation operator $\mathcal{S}_{\mathrm{evt}}(\cdot)$ must be re-applied to the corresponding linear intensity:
\begin{equation}
    \mathbf{E}_{\mathrm{ob}} = \mathcal{S}_{\mathrm{evt}}\left( \exp\left( c \int_{0}^{t} \mathbf{E}_{\mathrm{ideal}}(\tau) \, d\tau + L(0) \right) \right)~.
    \label{app:eq:full_forward_model}
\end{equation}
This completes our theoretical \textbf{forward model}. It reveals that the forward problem is governed by the intractable, stateful operator $\mathcal{S}_{\mathrm{evt}}$, making the inverse problem (de-flaring) analytically unsolvable and motivating our learning-based approach.

\vspace{0.1cm}
\subsection{~From Theoretical Model to a Practical Simulator}

While the full forward model (Eq.~\ref{app:eq:full_forward_model}) is too complex for direct implementation, the idealized model (Eq.~\ref{app:eq:hypo_stream_final_redef}) isolates the core physical principle: an \textbf{intensity-weighted fusion} of event streams. Our practical simulator is built directly upon this principle.

We implement the suppression mechanism as the \textbf{Probabilistic Non-Linear Event Suppression (PNL-ES)} operator, denoted by $\oplus$. This operator approximates the outcome of the complex integrate-and-fire process with a probabilistic gating mechanism, where the weights $w(t)$ are interpreted as the retention probabilities for incoming events. This provides an efficient and effective implementation of the intensity-weighted fusion. The process is summarized in Algorithm~\ref{app:alg:event_mixing}.

\begin{algorithm}[t]
\caption{PNL-ES Operator: Physics-Grounded Event Stream Fusion}
\label{app:alg:event_mixing}
\begin{algorithmic}[1]
\Statex \textbf{Input:} Event streams $\mathbf{E}_a$, $\mathbf{E}_b$; Intensity profiles $\mathbf{I}_a(t)$, $\mathbf{I}_b(t)$
\Statex \textbf{Output:} Fused event stream $\mathbf{E}_a \oplus \mathbf{E}_b$
\Statex
\State $\mathbf{E}_{\mathrm{fused}} \leftarrow \emptyset$
\State Calculate weights $w_a(t) = \mathbf{I}_a(t) / (\mathbf{I}_a(t) + \mathbf{I}_b(t))$ and $w_b(t) = 1 - w_a(t)$.
\For{each event $e_i$ in $\mathbf{E}_a$}
    \If{$\mathrm{random}() < w_a(e_i.t)$}
        \State Add $e_i$ to $\mathbf{E}_{\mathrm{fused}}$
    \EndIf
\EndFor
\For{each event $e_j$ in $\mathbf{E}_b$}
    \If{$\mathrm{random}() < w_b(e_j.t)$}
        \State Add $e_j$ to $\mathbf{E}_{\mathrm{fused}}$
    \EndIf
\EndFor
\State
\State \Return $\mathbf{E}_{\mathrm{fused}}$ sorted by timestamps
\end{algorithmic}
\end{algorithm}

Our simulator uses this operator to construct the paired training data. The scene is decomposed into three components:
\begin{itemize}
    \item \textbf{$\mathbf{E}_{\mathrm{bg}}$}: Events from the flare-free background scene.
    \item \textbf{$\mathbf{E}_{\textcolor{ev_red}{\mathrm{\mathbf{f}}}}$}: Events from the light source as captured by a \textbf{non-ideal} lens (with flare).
    \item \textbf{$\mathbf{E}_{\mathrm{light}}$}: Events from the same light source as captured by an \textbf{ideal} lens.
\end{itemize}

\vspace{0.1cm}
The network's task is to learn the mapping from the flare-corrupted observation $\mathbf{E}_{\mathrm{ob}}$ to the clean ground truth $\mathbf{E}_{\mathrm{gt}}$. These are synthesized as:
\begin{align}
\mathbf{E}_{\mathrm{ob}} &= \mathbf{E}_{\mathrm{bg}} \oplus \mathbf{E}_{\textcolor{ev_red}{\mathrm{\mathbf{f}}}}~, \label{app:eq:compose_obs_concise} \\
\mathbf{E}_{\mathrm{gt}} &= \mathbf{E}_{\mathrm{bg}} \oplus \mathbf{E}_{\mathrm{light}}~. \label{app:eq:compose_gt_concise}
\end{align}
This physically grounded synthesis pipeline enables the generation of a large-scale dataset for training a robust de-flaring network.

\vspace{0.1cm}
\subsection{~A Generalized Weighted Fusion Framework}
\label{app:generalized_framework}

The theoretical foundation of our model rests upon the linear superposition of irradiance. This two-component model represents a specific instance of a more general principle. We can extend this concept to a scenario where the final observed irradiance $\mathbf{I}_{\text{obs}}$ is a weighted summation of $N$ distinct, incoherent light sources or paths:
\begin{equation}
    \mathbf{I}_{\mathrm{obs}}(t) = \sum_{i=1}^{N} k_i \mathbf{I}_i(t)~.
    \label{eq:generalized_superposition}
\end{equation}
Here, $\mathbf{I}_i(t)$ is the irradiance of the $i$-th component, and $k_i$ is a time-invariant scalar coefficient representing the contribution factor of that component. These coefficients are immensely powerful as they can directly model physical properties. For example, when imaging through a glass window, $\mathbf{I}_1$ could be the transmitted scene and $\mathbf{I}_2$ the reflected scene, with $k_1$ and $k_2$ being the glass's transmissivity and reflectivity, respectively.

Following the same derivation logic, we can determine how this generalized superposition manifests in the event domain. We begin by translating Eq.~\ref{eq:generalized_superposition} into the logarithmic domain:
\begin{equation}
    L_{\mathrm{obs}}(t) = \log\left(\sum_{i=1}^{N} k_i \mathbf{I}_i(t)\right) = \log\left(\sum_{i=1}^{N} k_i e^{L_i(t)}\right)~.
\end{equation}
To find the resulting event stream, we compute the time derivative of $L_{\mathrm{obs}}(t)$. Applying the chain rule, we obtain:
\begin{equation}
\begin{split}
    \frac{d}{dt} L_{\mathrm{obs}}(t) &= \frac{1}{\sum_{j=1}^{N} k_j \mathbf{I}_j(t)} \sum_{i=1}^{N} k_i \frac{d}{dt}\mathbf{I}_i(t) \\
    &= \sum_{i=1}^{N} \frac{k_i \mathbf{I}_i(t)}{\sum_{j=1}^{N} k_j \mathbf{I}_j(t)} \frac{d}{dt}L_i(t)~.
\end{split}
\end{equation}
Here, we have used the identity $\frac{d}{dt}\mathbf{I}_i(t) = \mathbf{I}_i(t)\frac{d}{dt}L_i(t)$. This leads us to define a set of generalized dynamic weights, $w_i(t)$, for each component stream:
\begin{equation}
    w_i(t) = \frac{k_i \mathbf{I}_i(t)}{\sum_{j=1}^{N} k_j \mathbf{I}_j(t)}~,
    \label{eq:generalized_weights}
\end{equation}
where, by definition, $\sum_{i=1}^{N} w_i(t) = 1$. The weight $w_i(t)$ now represents the instantaneous fractional contribution of the \emph{effective} irradiance of component $i$ (\ie, $k_i \mathbf{I}_i(t)$) to the total observed irradiance.

By substituting the sensor model  for each of the $N$ components, we arrive at the generalized event stream fusion model:
\begin{equation}
    \mathbf{E}_{\mathrm{obs}}(t) + \frac{d}{dt}\varepsilon_{\mathrm{obs}}(t) = \sum_{i=1}^{N} w_i(t)\left[\mathbf{E}_i(t) + \frac{d}{dt}\varepsilon_i(t)\right]~.
    \label{eq:generalized_full_event_mixing}
\end{equation}
This equation is the central result of our generalization. It reveals that any scene composed of multiple weighted, additive light components results in an event stream that is a non-linear fusion of the individual source event streams, dynamically weighted by their relative effective intensities.

In the idealized scenario where sub-threshold drifts are negligible ($\frac{d}{dt}\varepsilon(t) \approx 0$), this complex interaction simplifies to a core principle:
\begin{equation}
    \mathbf{E}_{\mathrm{ideal}}(t) = \sum_{i=1}^{N} w_i(t)\mathbf{E}_i(t)~.
    \label{eq:generalized_hypo_stream_final}
\end{equation}
This generalized forward model provides a powerful theoretical tool for understanding and, more importantly, simulating a wide range of complex visual phenomena far beyond lens flare.

\vspace{0.1cm}
\subsection{~Additional Applications}
\label{app:potential_applications}

The generalized framework derived in Eq.~\ref{eq:generalized_hypo_stream_final}, with dynamic weights defined in Eq.~\ref{eq:generalized_weights}, extends far beyond the specific case of lens flare. The simulation-driven paradigm it enables can be applied to a broad spectrum of challenging event-based vision tasks that have historically suffered from a lack of ground-truth data. We detail three such applications below.

\vspace{0.2cm}
\noindent\textbf{Reflections and Transparency.} 
Imaging through semi-transparent surfaces, such as glass windows or water, is a classic computer vision problem. The observed scene can be physically modeled as a weighted fusion of two distinct components: the transmitted background radiance ($\mathbf{I}_{\mathrm{transmitted}}$) and the reflected foreground radiance ($\mathbf{I}_{\mathrm{reflected}}$). Under our framework, the observed irradiance $\mathbf{I}_{\mathrm{obs}}$ is formulated as:
\begin{equation}
    \mathbf{I}_{\mathrm{obs}} = k_T \mathbf{I}_{\mathrm{transmitted}} + k_R \mathbf{I}_{\mathrm{reflected}}~,
\end{equation}
where $k_T$ and $k_R$ represent the surface's transmissivity and reflectivity coefficients, respectively. By applying our generalized fusion model, we can synthesize realistic, paired training data (e.g., pure transmission vs. mixed observation) to train networks capable of separating these superimposed event streams, a task notoriously difficult with conventional methods.

\vspace{0.2cm}
\noindent\textbf{Handling Occlusions.} 
The framework naturally accommodates occlusion modeling. In the case of an opaque occluder, the model involves a background stream ($\mathbf{E}_{\mathrm{bg}}$) and an occluder stream ($\mathbf{E}_{\mathrm{occ}}$). The coefficients $k_i$ become spatially-dependent binary masks:
\begin{equation}
    k_{\mathrm{bg}}(x,y) = 
    \begin{cases} 
    1 & \text{if visible} \\
    0 & \text{if occluded}
    \end{cases}~, \quad
    k_{\mathrm{occ}}(x,y) = 1 - k_{\mathrm{bg}}(x,y)~.
\end{equation}
For semi-transparent occluders, such as smoke, fog, or sparse foliage, the coefficients generalize to continuous values $k_{\mathrm{bg}} \in (0,1)$, representing the degree of attenuation. This allows for the precise synthesis of complex occlusion scenarios, facilitating the training of robust networks for tasks like object tracking and recognition through partial or semi-transparent occlusion.

\vspace{0.2cm}
\noindent\textbf{Imaging in Participating Media.} 
Perhaps the most compelling application lies in modeling scenes within participating media, such as fog, heavy rain, or underwater environments. In these conditions, the light reaching the sensor is a complex integration of light attenuated from objects at various depths and light scattered by the medium itself. Our discrete superposition model (Eq.~\ref{eq:generalized_superposition}) can serve as an effective approximation of the volume rendering integral. Specifically, the contribution from a depth layer $z_i$ can be modeled with an attenuation coefficient:
\begin{equation}
    k_i \propto e^{-\alpha z_i}~,
\end{equation}
where $\alpha$ is the medium's extinction coefficient. By discretizing the scene depth into layers, our framework paves the way for creating physically-grounded simulators for adverse weather and underwater conditions, enabling the development of robust, event-based perception systems for extreme environments.

\vspace{0.2cm}
In summary, the fusion model initially developed for lens flare provides a unifying theoretical lens through which a multitude of complex event-based vision problems can be addressed. Its primary contribution is a principled mechanism for data synthesis, unlocking the potential of learning-based solutions for tasks that were previously intractable due to the physical impossibility of acquiring paired ground-truth data.
\clearpage\clearpage
\section{~E-DeflareNet: Network \& Codec Details}
\label{app:network_details}

This section provides further implementation details for our de-flaring network, \textbf{E-DeflareNet}, and the event-to-voxel codec used during the data representation learning process.

\vspace{0.1cm}
\subsection{~Network Architecture}

\textbf{E-DeflareNet} employs a custom \textbf{TrueResidualUNet3D} architecture following the residual learning principle: $\hat{\mathcal{V}}_{\mathrm{gt}} = \mathcal{V}(\mathbf{E}_{\mathrm{ob}}) + f_{\theta}(\mathcal{V}(\mathbf{E}_{\mathrm{ob}}))$. The backbone $f_{\theta}$ is adapted from \textbf{ResidualUNet3D} in the \textit{pytorch-3dunet} library~\cite{10.7554/eLife.57613}. A critical design choice is the zero-initialization of the final $1\times1\times1$ convolution, ensuring $f_{\theta}(\cdot) \approx 0$ at initialization. This forces an initial identity mapping, allowing the model to focus on learning the negative flare residual effectively.

\begin{figure}[b]
    \centering
    \vspace{-0.1cm}
    \includegraphics[width=\columnwidth]{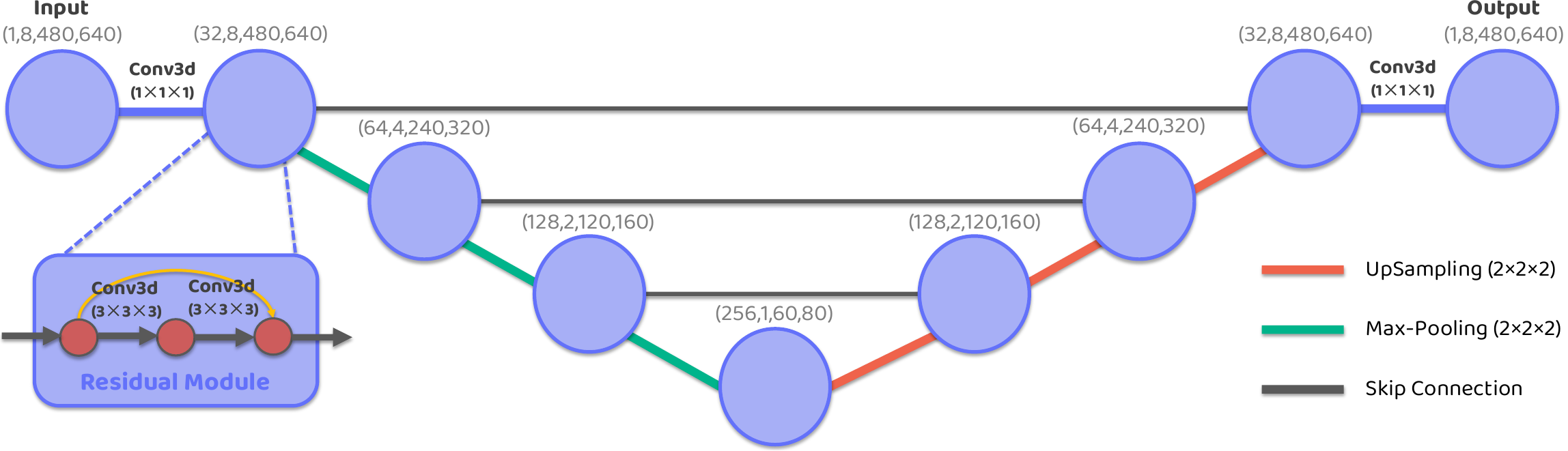}
    \vspace{-0.5cm}
    \caption{\textbf{E-DeflareNet Architecture.} The network is a standard $4$-level Residual 3D U-Net, featuring a symmetric encoder-decoder structure with skip connections to combine multi-scale features for high-fidelity event-based restoration.}
    \label{fig:network_architecture}
\end{figure}

\vspace{0.1cm}
Illustrated in Fig.~\ref{fig:network_architecture}, the network is a 4-level U-Net with $\sim$$7.07$M parameters. It accepts single-channel voxel grids of shape $(B, 1, 8, 480, 640)$. The encoder utilizes residual blocks and 3D max-pooling to downsample features, while the symmetric decoder uses transposed convolutions for upsampling. Skip connections concatenate encoder features with the decoder path, integrating multi-scale context for high-fidelity restoration.

\vspace{0.1cm}
\subsection{~Event-to-Voxel Encoding}

The encoding operator, $\mathcal{V}(\cdot)$, transforms an event stream $\mathbf{E} = \{e_k\}_{k=1}^N$ over a duration $T$ (20ms) into a voxel grid $\mathcal{V} \in \mathbb{R}^{B_\mathrm{bins} \times H \times W}$. Each event is a tuple $e_k = (t_k, x_k, y_k, p_k)$. The process is as follows:

\begin{enumerate}
    \item \textbf{Initialization:} A zero tensor $\mathcal{V}$ of size $(B_\mathrm{bins}, H, W)$ is created, where $B_\mathrm{bins}=8$ is the number of temporal bins.
    \item \textbf{Temporal Binning:} The duration $T$ is divided into $B_\mathrm{bins}$ uniform intervals, each of duration $\Delta t = T/B_\mathrm{bins}$. For each event $e_k$, its corresponding bin index $b_k$ is calculated as $b_k = \lfloor (t_k - t_0) / \Delta t \rfloor$, where $t_0$ is the timestamp of the first event.
    \item \textbf{Polarity Accumulation:} The polarity $p_k$ of each event is accumulated at the corresponding spatio-temporal cell. The value at cell $(b, y, x)$ is given by:
    \begin{equation}
        \mathcal{V}(b, y, x) = \sum\nolimits_{k: (x_k, y_k, b_k) = (x, y, b)} p_k~.
    \end{equation}
\end{enumerate}
This process is implemented using efficient, vectorized PyTorch operations for parallel processing.

\vspace{0.1cm}
\subsection{~Voxel-to-Event Decoding}

The decoding operator, $\mathcal{V}^{-1}(\cdot)$, converts a restored voxel grid $\hat{\mathcal{V}}_{\mathrm{gt}}$ back into a clean event stream $\hat{\mathbf{E}}_{\mathrm{gt}}$. This procedure inverts the encoding process:

\begin{enumerate}
    \item \textbf{Event Count Determination:} For each cell $(b, y, x)$, its floating-point value is rounded to the nearest integer, $n = \text{round}(\hat{\mathcal{V}}_{\mathrm{gt}}(b, y, x))$. The absolute value $|n|$ determines the number of events, and $\text{sgn}(n)$ determines their polarity.
    \item \textbf{Timestamp Generation:} For the $|n|$ events generated from a given bin $b$, each is assigned a random timestamp drawn from a uniform distribution within that bin's time interval: $t \sim \mathcal{U}(b \cdot \Delta t, (b+1) \cdot \Delta t)$. This restores the asynchronous nature of the events.
    \item \textbf{Stream Assembly:} All events generated from all cells are collected into a single list.
    \item \textbf{Chronological Sorting:} The entire list of events is sorted by timestamp to produce the final, valid event stream $\hat{\mathbf{E}}_{\mathrm{gt}}$.
\end{enumerate}
\clearpage\clearpage
\section{~Additional Implementation Details}
\label{app:experimental_details}

This section provides a comprehensive and detailed breakdown of our experimental methodology. We elaborate on the specific training protocols employed for \textbf{E-DeflareNet}, the rigorous configuration of baseline methods to ensure fair comparison, and the formal definitions of the multi-level evaluation metrics used to assess performance.

\vspace{0.1cm}
\subsection{~Training Details}
\label{app:training_details}
We implemented \textbf{E-DeflareNet} using the PyTorch framework. The network is optimized to learn the residual flare component by minimizing the Mean Squared Error (MSE) loss computed on the voxel grids. We employed the \textbf{Adam} optimizer with an initial learning rate of $2 \times 10^{-4}$ and a weight decay of $1 \times 10^{-5}$ for regularization. The learning rate was modulated using a \texttt{MultiStepLR} scheduler, which decayed the rate by a factor of $0.2$ at epochs $10$, $20$, and $30$.

Training was conducted on a single NVIDIA RTX 3090 GPU with a batch size of $2$. We utilized the training split of the E-Flare-2.7K dataset, which comprises $2{,}545$ paired samples. The final model was selected from the checkpoint saved at iteration $40{,}000$.

To ensure training stability and accommodate the unbounded nature of event voxel values (ranging from $-\infty$ to $+\infty$), the network employs an identity activation at the output layer rather than a sigmoid or softmax function. Crucially, when combined with the zero-initialization of the final convolution layer (as detailed in Sec.~\ref{app:network_details}), this configuration guarantees that the model initializes as a perfect identity mapping. This allows the optimizer to focus immediately on learning the subtractive flare interference without the need to correct initial random perturbations.

\vspace{0.1cm}
\subsection{~Baseline Methods}
\label{app:baselines_details}

This section details the specific configurations and necessary adaptations made to the baseline methods. Our goal was to ensure a fair and robust comparison by aligning the operating conditions of all methods while keeping the core algorithms of external baselines consistent with their original publications.

\subsubsection{EFR}
Event Frequency Removal (EFR) \cite{wang2022linear} is a linear comb filter primarily designed to remove periodic noise sources. For our experiments, we configured the filter to target a base frequency of $50$~Hz, utilizing the default primary feedback coefficient ($\rho_1 = 0.6$). However, a critical adaptation was required to facilitate a fair evaluation on our dataset. The original EFR implementation includes a warm-up phase that discards all events within the initial $22$~ms. In the context of our short-duration test sequences, this behavior would result in significant and unfair data loss. To address this, we developed a Python wrapper that preserves these initial warm-up events and re-appends them, unmodified, to the filtered output stream, ensuring that the evaluation is conducted on the complete temporal window.

\subsubsection{PFD-A \& PFD-B}
These baselines represent two distinct variants of a polarity-flip-based denoising algorithm proposed in \cite{shi2024polarity}. Following the recommendations provided by the authors, we evaluated two specific configurations:
\begin{itemize}
    \item \textbf{PFD-A} (\texttt{score\_select=1}): This variant is configured for general-purpose event denoising, utilizing spatio-temporal correlation to filter out isolated noise events.
    
    \item \textbf{PFD-B} (\texttt{score\_select=0}): This variant is specifically tailored for removing artifacts originating from light sources. By focusing on polarity flipping patterns, it serves as a stronger and more relevant baseline for the specific task of de-flaring.
\end{itemize}
For both methods, we adhered to the suggested parameters, using a temporal window of $20$~ms and a spatial neighborhood size of $3$. Our implementation utilizes a Python wrapper to directly invoke the original C++ executable, thereby ensuring full algorithmic consistency with the reported results.

\vspace{0.1cm}
\subsubsection{Voxel Transform}
The \textit{Voxel Transform} baseline serves as a rigorous control experiment designed to isolate the specific contribution of our neural network. Its primary purpose is to verify that the observed de-flaring effect originates exclusively from the learned restoration model, rather than from any implicit filtering artifacts that might be introduced by the event-to-voxel encoding and decoding process. The pipeline for this baseline encodes the input event stream into a voxel grid, which then \textbf{bypasses the E-DeflareNet entirely} and is immediately decoded back into an event stream. Effectively, this baseline represents a ``null" network where $f_{\theta}(\cdot) = 0$. Consequently, any performance improvement of our full model over this baseline is directly attributable to the network's learned capability to identify and remove flare artifacts.

\vspace{0.1cm}
\subsection{Evaluation Metrics \& Protocol}
\label{app:metrics_details}

To provide a holistic assessment of restoration quality, our evaluation employs a comprehensive suite of metrics operating at two distinct levels: the raw, asynchronous event level and the structured, volumetric voxel level. All reported scores are computed by averaging the results over $20$~ms segments of the event streams.

\vspace{0.1cm}
\subsubsection{Event-Level Evaluation Metrics}
These metrics treat the event streams as 4D point clouds in space-time $(t, x, y, p)$. Prior to metric computation, all four dimensions are normalized to a unified range of $[0, 100]$ to ensure balanced contribution from spatial and temporal errors.

\vspace{0.1cm}
\noindent\textbf{Chamfer Distance (CD).}
The Chamfer Distance measures the average distance from each event in the restored stream to its nearest neighbor in the ground-truth stream. We use a one-sided variant, as we do not want to penalize the correct removal of flare events while not adding noise. It is defined as:
\begin{equation}
    \text{CD}(\hat{\mathbf{E}}_{\mathrm{gt}}, \mathbf{E}_{\mathrm{gt}}) = \frac{1}{|\hat{\mathbf{E}}_{\mathrm{gt}}|} \sum_{e_i \in \hat{\mathbf{E}}_{\mathrm{gt}}} \min_{e_j \in \mathbf{E}_{\mathrm{gt}}} \| e_i - e_j \|_2~,
\end{equation}
where $\| \cdot \|_2$ denotes the Euclidean distance in the normalized 4D space. A lower CD value indicates better alignment and higher fidelity.

\vspace{0.1cm}
\noindent\textbf{Gaussian Distance (GD).}
The Gaussian Distance is a robust variant of the Chamfer Distance. It incorporates a Gaussian weighting kernel to the nearest-neighbor distances, which effectively suppresses the influence of outliers and focuses the metric on local alignment quality. It is defined as:
\begin{equation}
    \text{GD}(\hat{\mathbf{E}}_{\mathrm{gt}}, \mathbf{E}_{\mathrm{gt}}) = \frac{1}{|\hat{\mathbf{E}}_{\mathrm{gt}}|} \sum_{e_i \in \hat{\mathbf{E}}_{\mathrm{gt}}} \left( 1 - \exp\left(-\frac{d_i^2}{\sigma}\right) \right)~,
\end{equation}
where $d_i^2 = \min_{e_j \in \mathbf{E}_{\mathrm{gt}}} \| e_i - e_j \|_2^2$ is the squared distance, and the kernel width parameter $\sigma$ is set to 0.4. A lower GD score indicates superior performance.

\vspace{0.1cm}
\subsubsection{Voxel-Level Evaluation Metrics}
These metrics evaluate performance on the discretized voxel grid representation. They compare the restored voxel grid $\hat{\mathcal{V}}_{\mathrm{gt}}$ against the ground-truth grid $\mathcal{V}_{\mathrm{gt}} \in \mathbb{R}^{B_\mathrm{bins} \times H \times W}$, where $B_\mathrm{bins}=8$ represents the number of temporal bins.

\vspace{0.1cm}
\noindent\textbf{Mean Squared Error (MSE).}
MSE serves as a strict, voxel-wise metric that computes the mean squared difference between the two voxel grids. It penalizes any deviation in event density without applying spatial tolerance:
\begin{equation}
    \text{MSE}(\hat{\mathcal{V}}_{\mathrm{gt}}, \mathcal{V}_{\mathrm{gt}}) = \frac{1}{B_\mathrm{bins} \cdot H \cdot W} \sum_{b,y,x} \left( \hat{\mathcal{V}}_{\mathrm{gt}}(b,y,x) - \mathcal{V}_{\mathrm{gt}}(b,y,x) \right)^2~.
\end{equation}

\vspace{0.1cm}
\noindent\textbf{Pooled Mean Squared Error (PMSE).}
Recognizing that exact pixel-level alignment can be overly harsh for event data, PMSE introduces spatial tolerance by applying a 2D average pooling operation prior to the MSE calculation. The voxel grids are first separated by polarity into a $\mathcal{V}_{\mathrm{pol}} \in \mathbb{R}^{2 \times B_\mathrm{bins} \times H \times W}$ representation. Then, a pooling operation $\text{Pool}_k$ with kernel size $k \times k$ is applied. PMSE is defined as:
\begin{equation}
    \text{PMSE}_k(\hat{\mathcal{V}}_{\mathrm{gt}}, \mathcal{V}_{\mathrm{gt}}) = \text{MSE}(\text{Pool}_k(\hat{\mathcal{V}}_{\mathrm{gt, pol}}), \text{Pool}_k(\mathcal{V}_{\mathrm{gt, pol}}))~.
\end{equation}
We report PMSE with pooling sizes of $k=2$ and $k=4$ (denoted as PMSE-2 and PMSE-4) to evaluate structural consistency at different scales.

\vspace{0.1cm}
\noindent\textbf{F1-Score Variants.}
These metrics focus on the structural similarity of the event distribution. They operate by binarizing the polarity-separated voxel grids with a threshold of $\theta = 0.001$ and computing the F1-score. We employ three variants with decreasing levels of strictness:
\begin{itemize}
    \item \textbf{Raw F1 (R-F1):} This score is computed directly on the full-dimensional grids $\mathcal{V}_{\mathrm{pol}} \in \mathbb{R}^{2 \times B_\mathrm{bins} \times H \times W}$. It is the strictest variant, requiring precise alignment in space, time, and polarity.
    
    \item \textbf{Temporal F1 (T-F1):} For this variant, the temporal dimension ($B_\mathrm{bins}$) is collapsed by summation before binarization. This metric is more lenient towards small temporal misalignments within the $20$~ms window.
    
    \item \textbf{Temporal-Polarity F1 (TP-F1):} This is the most relaxed variant, where both the temporal and polarity dimensions are collapsed. It evaluates the match of the overall spatial event distribution, disregarding timing and polarity.
\end{itemize}
For all F1 variants, a higher score indicates better structural recovery.
\clearpage\clearpage
\section{~Additional Experimental Results}
\label{app:more_results}

In this section, we present an extended collection of qualitative results to provide a more comprehensive demonstration of the effectiveness and robustness of our \textbf{E-Deflare} framework. These visualizations cover a broader spectrum of challenging scenarios, ranging from standard driving conditions to edge cases. 

\vspace{0.1cm}
We specifically highlight two distinct categories of flare artifacts to analyze model performance: typical urban interference patterns and extreme, atypical scenarios that test the limits of generalization.

\vspace{0.1cm}
\subsection{~Robustness in Typical Urban Scenarios} 

\begin{figure}[t]
    \centering
    \includegraphics[width=\textwidth]{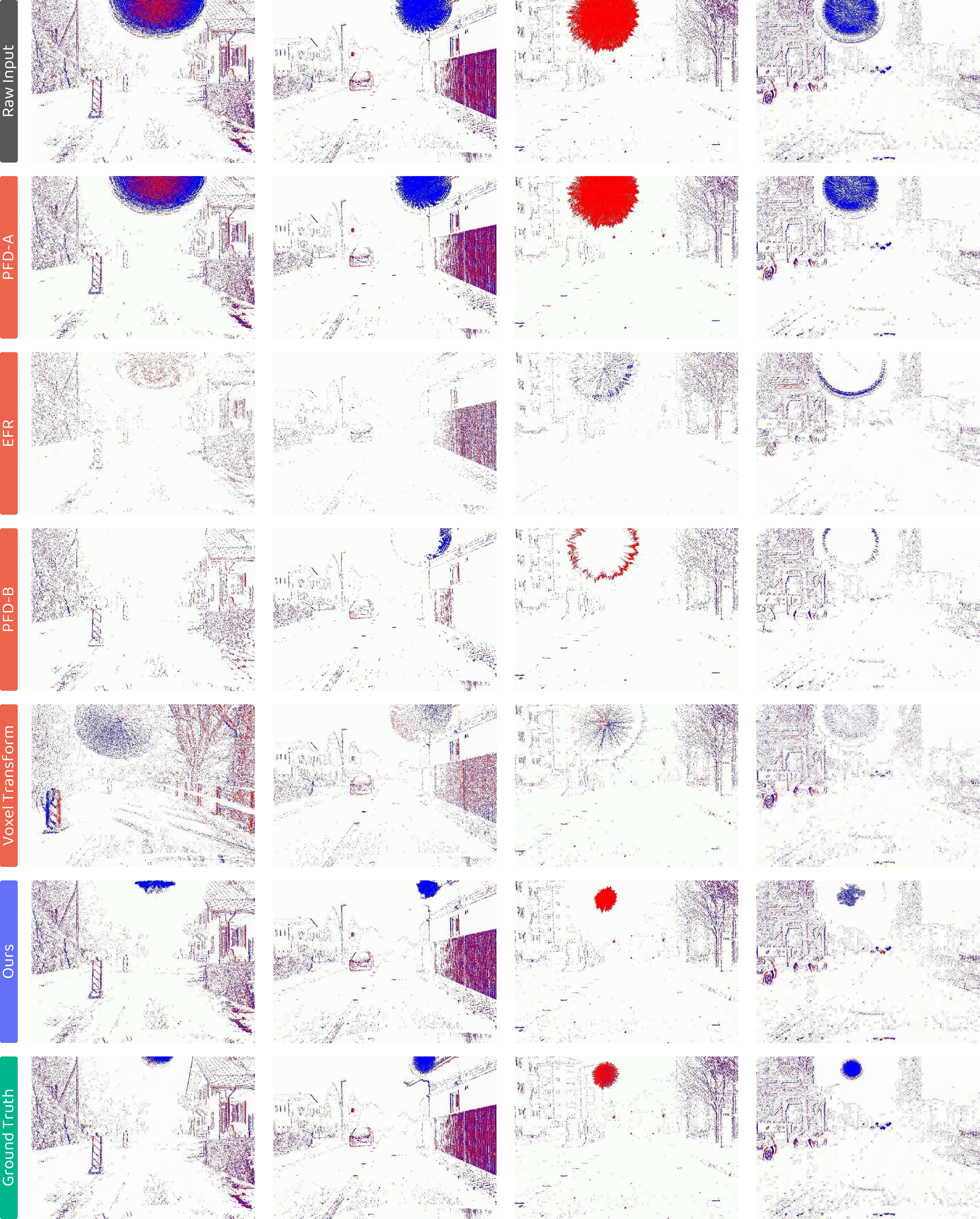}
    \vspace{-0.55cm}
    \caption{\textbf{Additional qualitative assessments on E-Flare2.7K}. Each row shows the output of a different method applied to the same flare-corrupted input. The bottom row displays the ground truth for reference. Best viewed in colors and zoomed-in for details.}
    \label{fig:simu_R_supp_1}
\end{figure}

\begin{figure}[t]
    \centering
    \includegraphics[width=\textwidth]{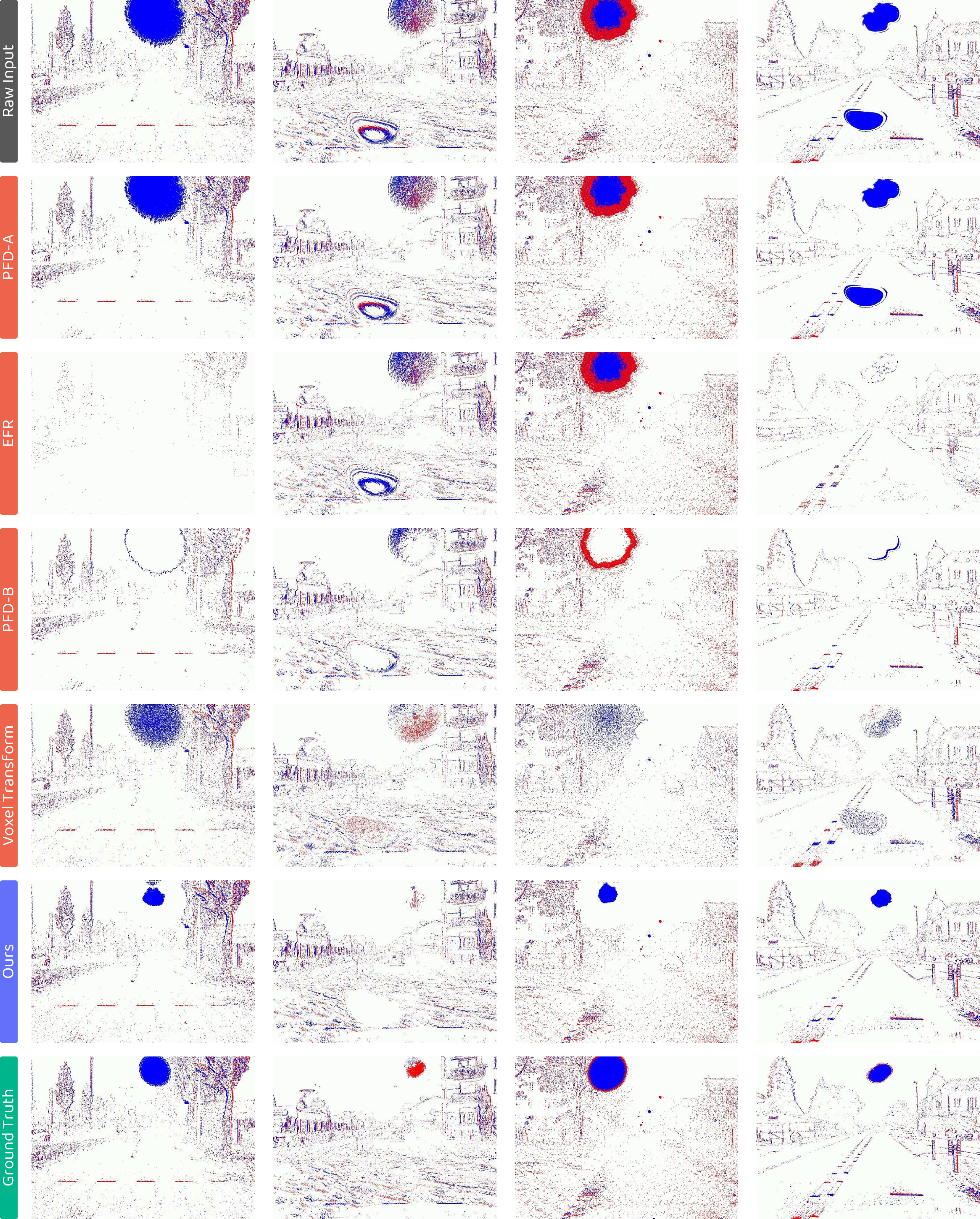}
    \vspace{-0.55cm}
    \caption{\textbf{Additional qualitative assessments on E-Flare2.7K}. Each row shows the output of a different method applied to the same flare-corrupted input. The bottom row displays the ground truth for reference. Best viewed in colors and zoomed-in for details.}
    \label{fig:simu_R_supp_2}
\end{figure}

Figure~\ref{fig:simu_R_supp_1} and Figure~\ref{fig:simu_R_supp_2} illustrate scenarios that are characteristic of urban night driving environments. These scenes are dominated by strong, periodic interference originating from streetlights -- a distribution that closely mirrors the real-world conditions found in the DSEC dataset \cite{gehrig2021dsec}. In these common yet challenging situations, the superiority of \textbf{E-DeflareNet} over competing approaches is evident. 

\vspace{0.1cm}
Existing baseline methods struggle to maintain a balance between artifact removal and signal preservation. Aggressive filtering techniques often incur severe collateral damage, erroneously suppressing a significant amount of valid background events alongside the flare, which leads to a loss of scene detail. Conversely, general denoising methods like PFD-A, which rely primarily on local spatio-temporal consistency, often fail to distinguish the flare from the scene. They tend to misinterpret the large-scale, structured nature of lens flare as a valid signal, consequently leaving the artifacts largely intact. 

\vspace{0.1cm}
In contrast, our method accurately parses the underlying structure of the flare, achieving precise removal of the interference while robustly maintaining the integrity and density of the valid background event stream.

\vspace{0.1cm}
\subsection{~Generalization to Extreme \& Static Cases} 

\begin{figure}[t]
    \centering
    \includegraphics[width=\textwidth]{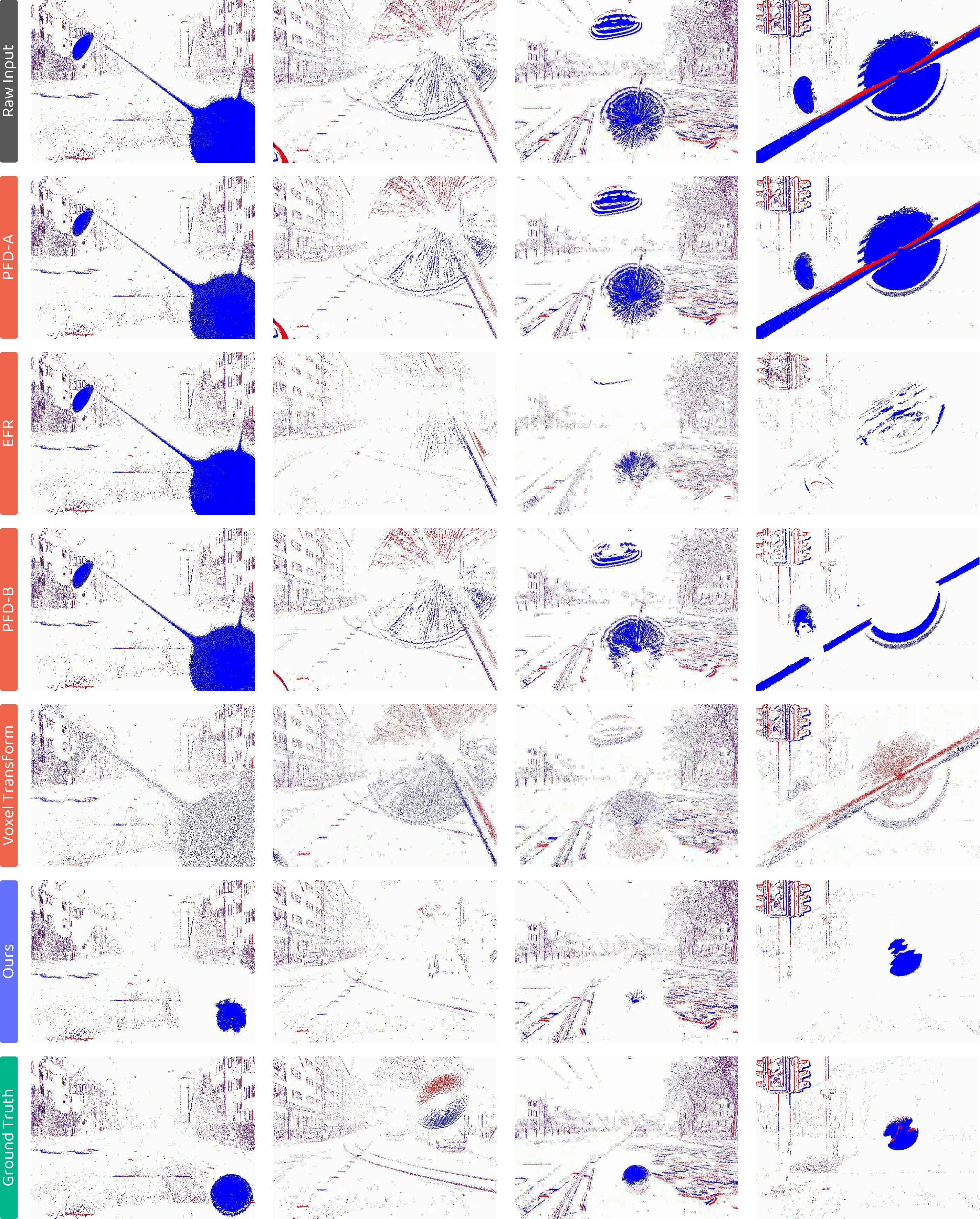}
    \vspace{-0.55cm}
    \caption{\textbf{Additional qualitative assessments on E-Flare2.7K}. Each row shows the output of a different method applied to the same flare-corrupted input. The bottom row displays the ground truth for reference. Best viewed in colors and zoomed-in for details.}
    \label{fig:simu_R_supp_3}
\end{figure}

Figure~\ref{fig:simu_R_supp_3} investigates the generalization capability of our model under stress tests that push the boundaries of standard datasets. These examples feature extreme flare instances resulting from light sources positioned in extreme proximity to the lens, generating massive artifacts with atypical, non-standard spatial shapes. 

\vspace{0.1cm}
Crucially, the examples presented in the \textbf{second and third columns} introduce a distinct challenge: they feature \textbf{non-flickering (static) light sources}, resembling continuous sunlight or stable DC lighting. This specific condition represents a critical failure mode for frequency-based baselines (such as EFR). Since these methods rely heavily on identifying specific temporal periodicity cues to detect interference, they become ineffective when the flare does not exhibit flickering characteristics. Consequently, such baselines often fail completely in these static cases. 

\vspace{0.1cm}
However, \textbf{E-DeflareNet} remains highly stable and effective. Because our physics-driven approach learns to recognize the intrinsic spatio-temporal morphology of the flare rather than relying solely on temporal frequency, it successfully suppresses these massive, static artifacts. This demonstrates a robust ability to generalize well beyond the typical flickering streetlight patterns encountered during training.

\vspace{0.1cm}
\subsection{~Video Demos}

Given that event data is inherently temporal, static images may not fully capture the dynamic quality of the restoration. Therefore, we provide three video demos to showcase the temporal consistency and stability of the de-flaring effect. 

\vspace{0.1cm}
These demos are attached to the supplementary material and include detailed qualitative comparisons on the \textbf{E-Flare-2.7K} (\texttt{demo1.mp4}), \textbf{E-Flare-R} (\texttt{demo2.mp4}), and \textbf{DSEC-Flare} (\texttt{demo3.mp4}) datasets, respectively.

\clearpage\clearpage
\section{~Broader Impact \& Limitations}
\label{app:impact_limitations}

\vspace{0.1cm}
\subsection{~Broader Impact}

While this work focuses on lens flare removal, the theoretical framework and simulation-driven paradigm we propose have far-reaching implications for the field of neuromorphic vision.

\vspace{0.1cm}
\noindent\textbf{A Unified Framework for Additive Interference.}
Our derivation of the non-linear suppression mechanism provides a foundational tool for modeling any scenario where multiple light transport paths interact. As detailed in Sec.~\ref{app:potential_applications}, this extends well beyond lens flare. It offers a mathematically grounded blueprint for synthesizing data and training restoration networks for a wide class of previously intractable problems, including:

\vspace{0.1cm}
\begin{itemize}
    \item \emph{Reflections and Transparency:} Separating transmitted and reflected event streams in glass or water surfaces.

    \vspace{0.1cm}
    \item \emph{Participating Media:} Modeling signal attenuation and scattering for perception in fog, rain, or underwater environments.

    \vspace{0.1cm}
    \item \emph{Partial Occlusions:} Handling semi-transparent obstacles like smoke or foliage.
\end{itemize}

\vspace{0.1cm}
\noindent\textbf{Catalyzing Cross-Modal Synergy.}
Our work also hints at the immense potential of fusing event data with standard frame-based (RGB) modalities. 

\vspace{0.1cm}
\begin{itemize}
    \item \emph{RGB-aided Event Restoration:} High-resolution RGB textures could serve as priors to guide the separation of flare from background events, improving the restoration of fine structural details that pure event-based methods might miss.

    \vspace{0.1cm}
    \item \emph{Event-guided Image Enhancement:} Conversely, the high temporal resolution of event cameras makes them ideal for detecting the rapid onset and trajectory of dynamic flares. Our de-flaring capability could act as a robust attention mechanism, helping RGB image signal processors (ISPs) better localize and remove flare artifacts in conventional video.
\end{itemize}

\vspace{0.1cm}
\subsection{~Societal Influence}
As a low-level enhancement module, \textbf{E-Deflare} poses no direct negative societal risks like labor displacement. Instead, it primarily benefits public safety by mitigating sensor blinding in autonomous vehicles during nighttime driving. Furthermore, our method adheres to privacy norms by recovering dynamic structures rather than sensitive static identities, maintaining the inherent privacy-preserving advantages of event cameras.

\vspace{0.1cm}
\subsection{~Potential Limitations}

Despite the promising results, our current framework has limitations. First, the scale of our training data and model parameters is relatively modest; scaling these up could further enhance restoration fidelity, particularly in recovering background events under extreme suppression. Second, the computational cost of the 3D convolutional backbone poses a challenge for resource-constrained edge devices. Future work will focus on model compression and lightweight architectures to ensure real-time processing efficiency.

\section{~Public Resource Used}

\subsection{~Public Datasets Used}
\begin{itemize}
    \item DSEC\footnote{\url{https://dsec.ifi.uzh.ch}} \dotfill CC BY-SA 4.0 License

    \item     Flare7K++\footnote{\url{https://github.com/ykdai/Flare7K}} \dotfill S-Lab License 1.0
\end{itemize}

\subsection{~Public Implementation Used}
\begin{itemize}
    \item SPADE-E2VID\footnote{\url{https://github.com/RodrigoGantier/SPADE_E2VID}}\dotfill Publicly Available

    \item Gaussian Splatting\footnote{\url{https://github.com/graphdeco-inria/gaussian-splatting}} \dotfill Inria / MPII License (Non-Commercial)

    \item pytorch-3dunet\footnote{\url{https://github.com/wolny/pytorch-3dunet}} \dotfill MIT License
    
    \item Flare Mania (Shadertoy)\footnote{\url{https://www.shadertoy.com/view/lsBGDK}} \dotfill CC BY-NC-SA 3.0
\end{itemize}

\end{document}